\documentclass[english, aps, prl, reprint, superscriptaddress, longbibliography]{revtex4-1}

\usepackage[T1]{fontenc}
\usepackage[latin9]{inputenc}
\usepackage{graphicx}
\usepackage{amsmath}
\usepackage{amssymb}
\usepackage{babel}
\usepackage{textcomp}
\usepackage{color}
\usepackage[normalem]{ulem}
\usepackage{cancel}

\usepackage{bm}
\newcommand{\configvec}{\bm}
\newcommand{\sitevec}{\vec}

\newcommand{\Tr}{\ensuremath{\mathrm{Tr}}}
\newcommand{\rmi}{\ensuremath{\mathrm{i}}}
\newcommand{\rmd}{\ensuremath{\mathrm{d}}}

\begin{document}

\title{Space of Functions Computed by Deep-Layered Machines }

\author{Alexander Mozeika}
\email{alexander.mozeika@kcl.ac.uk}
\affiliation{London Institute for Mathematical Sciences, London, W1K 2XF, United Kingdom}

\author{Bo Li}
\email{b.li10@aston.ac.uk}
\affiliation{Nonlinearity and Complexity Research Group, Aston University, Birmingham, B4 7ET, United Kingdom}

\author{David Saad}
\email{d.saad@aston.ac.uk}
\affiliation{Nonlinearity and Complexity Research Group, Aston University, Birmingham, B4 7ET, United Kingdom}

\begin{abstract}
We study the space of functions computed by random-layered machines, including deep neural networks and Boolean circuits. Investigating the distribution of Boolean functions computed on the recurrent and layer-dependent architectures, we find that it is the same in both models. Depending on the initial conditions and computing elements used, we characterize the space of functions computed at the large depth limit and show that the macroscopic entropy of Boolean functions is either monotonically increasing or decreasing with the growing depth.
\end{abstract}

\maketitle

Deep-layered machines comprise multiple consecutive layers of basic computing elements aimed at representing an arbitrary function, where the first and final layers represent its input and output arguments, respectively. Notable examples include deep neural networks (DNNs) composed of perceptrons~\cite{LeCun2015} and Boolean circuits constructed from logical gates~\cite{ODonnell2014}. Being universal approximators~\cite{Hornik1989, Poole2016}, DNNs have been successfully employed in different machine learning applications~\cite{LeCun2015}. Similarly, Boolean circuits can compute any Boolean function even when constructed from a single gate~\cite{Nisan2008}.

While the majority of DNN research focuses on their application in carrying out various learning tasks, it is equally important to establish the space of functions they typically represent for a given architecture and the computing elements used. One way to address such a generic study is to consider a random ensemble of DNNs. The study of random neural networks using methods of statistical physics has played an important role in understanding their \emph{typical} properties for storage capacity and generalization ability~\cite{Engel2001, Saad1998} and properties of energy-based~\cite{Agliari2012, Huang2015, Gabrie2015, Mezard2017, Barra2018} and associative memory models~\cite{Hopfield1982, Hertz1991}, as well as the links between energy-based models and feed-forward layered machines~\cite{Hopfield1987}.
In parallel, there have been theoretical studies within the computer science community of the range of Boolean functions generated by random Boolean circuits~\cite{Savicky1990, Brodsky2005}. Both the DNNs and the Boolean circuits share common basic properties.

Characterizing the space of functions computed by random-layered machines is of great importance, since it sheds light on their approximation and generalization properties.  However, it is also highly challenging due to the inherent recursiveness of computation and randomness in their architecture and/or computing elements.  Existing theoretical studies of the function space of deep-layered machines are mostly based on the mean field approach, which allows for a sensitivity analysis of the functions realized by deep-layered machines due to input or parameter perturbations~\cite{Poole2016, Lee2018, BoLi2018, BoLi2020}. 

To gain a complete and detailed understanding of the function space, we develop a path-integral formalism that directly examines \emph{individual functions} computed. This is carried out by processing 
\emph{all possible input configurations simultaneously and the corresponding outputs.}
For simplicity, we always consider Boolean functions with binary input and output variables.

The main contribution of this Letter is in providing a detailed understanding of the distribution of Boolean functions computed at each layer. It points to the equivalence between recurrent and layer-dependent  architectures, and consequently to the potential significant reduction in the number of trained free variables. Additionally, the complexity of Boolean functions implemented measured by their entropy, which depends on the number of layers and computing elements used, exhibits a rapid simplification when rectified linear unit (ReLU) components are employed, which arguably explains their generalization successes.

\textit{Framework.}\textendash\textendash 
The layered machines considered consist of $L+1$ layers, each with $N$ nodes. Node $i$ at layer $l$ is connected to the set of nodes $\{i_{1},i_{2},...,i_{k}\}$ of layer $l-1$; its activity is determined by the gate $\alpha_{i}^{l}$, computing a function of $k$ inputs, according to the propagation rule
\begin{equation}
P(S_{i}^{l}|\sitevec{S}^{l-1})=\delta\big[ S_{i}^{l},\alpha_{i}^{l}(S_{i_{1}}^{l-1},S_{i_{2}}^{l-1},...,S_{i_{k}}^{l-1})\big],\label{eq:prop}
\end{equation}
where $\delta$ is the  Dirac or Kronecker delta function, depending on the domain of $S_{i}^{l}$. The probabilistic form of Eq.~(\ref{eq:prop}) adopted here is convenient for the generating functional analysis and inclusion of noise~\cite{Mozeika2009, BoLi2018}. We primarily consider two structures here: (i) densely connected models where $k=N$ and node $i$ is connected to all nodes from the previous layer\textendash\textendash one such example is the fully connected neural network with $S_{i}^{l}=\alpha^{l}(H_{i}^{l})$,  where $H_{i}^{l}=\sum_{j=1}^{N}W_{j}^{l}S_{j}^{l-1}/\sqrt{N}+b_{i}^{l}$ is the preactivation field
and $\alpha^{l}$ is the activation function at layer $l$, (we will mainly focus on the case $b_{i}^{l}=0$; the effect of nonzero bias is discussed in~\cite{Mozeika2020sup}); (ii) sparsely connected models where $k\in O(N^{0})$\textendash\textendash examples include the sparse neural networks and layered Boolean circuits where $\alpha_{i}^{l}$ is a Boolean gate with $k$ inputs, e.g., majority gate.

Consider a binary input vector $\sitevec{s}=(s_{1},\ldots,s_{n})\in\{-1,1\}^{n}$, which is fed to the initial layer $l=0$. To accommodate a broader set of functions, we also consider an augmented input vector, e.g., (i) $\sitevec{S}^{I}=(\sitevec{s}, 1)$, which is equivalent to adding a bias variable in the context of neural networks; (ii) $\sitevec{S}^{I} = (\sitevec{s}, -\sitevec{s}, 1, -1)$, which has been used to construct all Boolean functions~\cite{Savicky1990}. Each node $i$ at layer $0$ points to a randomly chosen element of $\sitevec{S}^{I}$ such that
\begin{equation}
P^{0}(\sitevec{S}^{0}|\sitevec{s})=\prod_{i=1}^{N}P^{0}\big[ S_{i}^{0}|S_{n_{i}}^{I}(\sitevec{s}) \big]=\prod_{i=1}^{N}\delta \big[ S_{i}^{0},S_{n_{i}}^{I}(\sitevec{s}) \big], \label{eq:P_S0_given_SI}
\end{equation}
where $n_{i}=1,...,|\sitevec{S}^{I}|$ is an index chosen from the flat distribution $P(n_{i})=1/|\sitevec{S}^{I}|$. 
\begin{figure}
    \centering
    \hspace{-7mm}
    \includegraphics[scale=0.56]{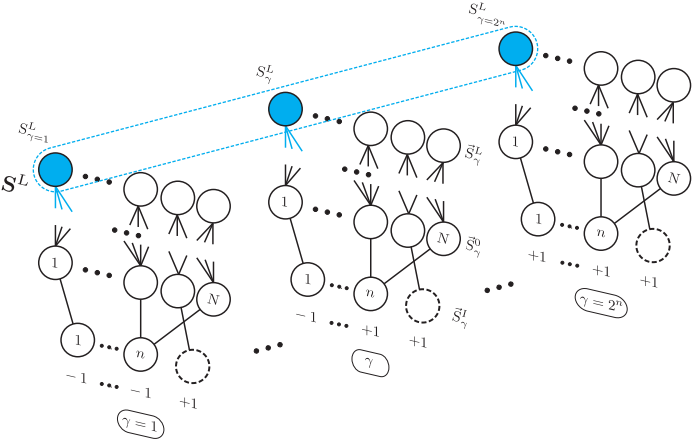}
    \caption{A deep-layered machine computing all possible $2^{n}$ inputs. The direction of computation is from bottom to top.  The binary string $\configvec{S}^{L}\in\{-1,1\}^{2^n}$ represents the Boolean function computed on the blue nodes of the output layer $L$. The augmented vector $\sitevec{S}^{I} = (\sitevec{s}, 1)$ is used as an example of input here. The constant $1$ is represented by the dashed circle.}
    \label{fig:illustrate}
\end{figure}

The computation of the layered machine is governed by the propagator $P(\sitevec{S}^{L}|\sitevec{s})=\sum_{\sitevec{S}^{L-1} \cdots \sitevec{S}^{0}}P(\sitevec{S}^{0}|\sitevec{s})\prod_{l=1}^{L}P(\sitevec{S}^{l}|\sitevec{S}^{l-1})$, where each node at layer $L$ computes a Boolean function $\{-1,1\}^{n}\to\{-1,1\}$. When the gates $\alpha_{i}^{l}$ or the network topology are \emph{random}, then the layered machine can be viewed as a disordered dynamical system with \emph{quenched} disorder~\cite{Mozeika2009,BoLi2018}. To probe the functions being computed, we consider the simultaneous layer propagation of \emph{all} possible inputs $\sitevec{s}_{\gamma}\in\{-1,1\}^n$, labeled by $\gamma=1,\ldots,2^{n}$ governed by the product propagator $\prod_{\gamma=1}^{2^{n}}P(\sitevec{S}_{\gamma}^{L}|\sitevec{s}_{\gamma})$. The binary string $\configvec{S}_{i}^{L}\in\{-1,1\}^{2^n}$ represents the Boolean function computed at node $i$ at layer $L$, as illustrated in Fig.~\ref{fig:illustrate}. Note that we use the vector notation $\sitevec{S}^{l}=(S_{1}^{l},...,S_{i}^{l},...,S_{N}^{l})$  and $\configvec{S}_{i}^{l}=(S_{i,1}^{l},...,S_{i,\gamma}^{l},...,S_{i,2^{n}}^{l})$ to represent the states and functions, respectively.
Using the above formalism, the distribution of Boolean functions $\configvec{f}$ computed on the final layer is given by
\begin{eqnarray}
P_{N}^{L}(\configvec{f})&=&\frac{1}{N}\sum_{i=1}^{N}\left\langle\prod_{\gamma=1}^{2^{n}}\delta\left(f_\gamma, S_{i,\gamma}^{L}\right)\right\rangle,\label{def:Boolean-F-density}
\end{eqnarray}
where components of $\configvec{f}$ satisfy $f_{\gamma} = f(\sitevec{s}_{\gamma})$, and angular brackets represent the average generated by $\prod_{\gamma=1}^{2^{n}}P(\sitevec{S}_{\gamma}^{L}|\sitevec{s}_{\gamma})$.
To compute $P_N^L(\configvec{f})$ and averages of other macroscopic \emph{observables}, which are expected to be self-averaging for $N\to\infty$~\cite{Mezard1987}, we introduce the disorder-averaged generating functional (GF) $\overline{\Gamma[\{\psi_{i,\gamma}^{l}\}]}=\overline{ \sum_{\{\sitevec{S}_{\gamma}^{l}\}} \prod_{\gamma} P(\sitevec{S}_{\gamma}^{0}|\sitevec{S}_{\gamma}^{I})\prod_{l}P(\sitevec{S}_{\gamma}^{l}|\sitevec{S}_{\gamma}^{l-1})e^{-\mathrm{i}\sum_{i}\psi_{i,\gamma}^{l}S_{i,\gamma}^{l}} }$, where the overline denotes an average over the quenched disorder. To keep the presentation concise, we outline the GF formalism only for DNNs in the following and refer the reader to~\cite{Mozeika2020sup} for the details of the derivation used in Boolean circuits.

\textit{Layer-dependent and recurrent architectures.}\textendash\textendash
We focus on two different architectures: layer-dependent  architectures, where the gates and/or connections are different from layer to layer, and recurrent, where the gates and connections are shared across all layers. Both architectures represent feed-forward machines that implement input-output mappings.

Specifically, we assume that the weights $W_{ij}^{l}$ in fully connected DNNs with layer-dependent  architectures are independent Gaussian random variables sampled from $\mathcal{N}(0,\sigma^{2})$. In DNNs with recurrent architectures, the weights are sampled once and are shared among layers, i.e. $W_{ij}^{l+1}=W_{ij}^{l}$.
We apply the sign activation function in the final layer, i.e. $\alpha^{L}(h_{i}^{L})=\mathrm{sgn}(h_{i}^{L})$, to ensure that the output of the DNN is Boolean.

We first outline the derivation for fully connected recurrent architectures.
It is sufficient to characterize the disorder-averaged GF by introducing cross-layer overlaps $q_{\gamma\gamma'}^{l,l'}=(1/N)\sum_{i}\overline{\langle S_{i,\gamma}^{l}S_{i,\gamma'}^{l'}\rangle}$ as order parameters and the corresponding conjugate order parameter $Q_{\gamma\gamma'}^{l,l'}$, which leads to a saddle-point integral $\overline{\Gamma}=\int\{\mathrm{d}\configvec{q}\,\mathrm{d}\configvec{Q}\}\mathrm{e}^{N\Psi[\configvec{q},\configvec{Q}]}$ with the potential~\cite{Mozeika2020sup}
\begin{eqnarray}
&&\Psi\!= 
 \rmi \Tr\left\{\configvec{q}\,\configvec{Q} \right\} 
 +\sum_{m=1}^{|\sitevec{S}^{I}|}P(m)\ln \sum_{\configvec{S}}\!\int\!\rmd\configvec{H} \mathcal{M}_{m}[\configvec{H},\configvec{S}],
\end{eqnarray}
where $\mathcal{M}_{m}[\configvec{H}, \configvec{S}]$ is an effective single-site measure 
\begin{eqnarray}
\mathcal{M}_{m} &=& \mathrm{e}^{-\rmi\sum_{l,\gamma}\psi_{\gamma}^{l}S_{\gamma}^{l}-\rmi\sum_{ll',\gamma\gamma'}Q_{\gamma\gamma'}^{l,l'}S_{\gamma}^{l}S_{\gamma'}^{l'}} \label{eq:def_Mm_recurrent} \\
 && \times\mathcal{N}\big(\configvec{H}\vert\configvec{0},\configvec{C}\big)\prod_{\gamma=1}^{2^{n}}P^{0}(S_{\gamma}^{0}|S_{m,\gamma}^{I})\prod_{l=1}^{L}\delta\big[S_{\gamma}^{l},\alpha^{l}(h_{\gamma}^{l})\big]. \nonumber
\end{eqnarray}
Due to weight sharing, the preactivation fields $\configvec{H}=(\configvec{h}^1,\ldots,\configvec{h}^L)$, where $\configvec{h}^l\in \mathbb{R}^{2^n}$, are governed by the Gaussian distribution $\mathcal{N}\left(\configvec{H} \vert \configvec{0}, \configvec{C}\right)$ and correlated across layers with covariance $[\configvec{C}\,]_{\gamma\gamma'}^{l,l'}=\sigma^{2}q_{\gamma\gamma'}^{l-1,l'-1}$. Setting $\psi_{\gamma}^{l}$ to zero and differentiating $\Psi$ with respect to $\{q_{\gamma\gamma'}^{l,l'},Q_{\gamma\gamma'}^{l,l'}\}$ yields the saddle point of the potential $\Psi$ dominating $\overline{\Gamma}$ for $N\rightarrow\infty$, at which the conjugate order parameters $Q_{\gamma\gamma'}^{l,l'}$ vanish~\cite{Mozeika2020sup}, leading to
\begin{equation}
q_{\gamma\gamma'}^{l,l'}=\begin{cases}
\sum_{m} P(m)\big\langle S_{\gamma}^{l}S_{\gamma'}^{0}\big\rangle_{\mathcal{M}_{m}}, & l'=0\\
\int\rmd\configvec{H}\, \alpha^{l}(h_{\gamma}^{l})\,\alpha^{l'}(h_{\gamma'}^{l'})\,  \mathcal{N}\big(\configvec{H}|\configvec{0}, \configvec{C}\big). & l'>0
\end{cases}\label{eq:saddle_q_recurrent}
\end{equation}
Notice that in the above Gaussian average, all preactivation fields but the pair $\{ h^l_{\gamma}, h^{l'}_{\gamma'} \}$ can be integrated out, reducing it to a tractable two-dimensional integral.

The GF analysis can be performed similarly for layer-dependent  architectures. Here the result has the same form as Eq.~(\ref{eq:saddle_q_recurrent}) with $q_{\gamma\gamma'}^{l,l'}=\delta_{l,l'} q_{\gamma\gamma'}^{l,l'}$, i.e. the overlaps between different layers are absent~\cite{Mozeika2020sup}, implying $[\configvec{C}\,]_{\gamma\gamma'}^{l,l'}=\sigma^{2}\delta_{l-1,l^\prime-1}q_{\gamma\gamma'}^{l-1,l'-1}$ for the covariances of preactivation fields. In this case, we denote the equal-layer covariance matrix as $\configvec{c}^{l} := \configvec{C}^{l,l}$.

We remark that the behavior of DNNs with layer-dependent  architectures in the limit of $N\to\infty$ can also be studied by mapping to Gaussian processes~\cite{Poole2016,Lee2018,Yang2019}. However, it is not clear if such analysis is possible in the highly correlated recurrent case while the GF or path-integral framework is still applicable~\cite{Coolen2001, Toyoizumi2015, Crisanti2018}.

Marginalizing the effective single-site measure in Eq.~(\ref{eq:def_Mm_recurrent}) gives rise to the distribution of Boolean functions $\configvec{f}\in\{-1,1\}^{2^n}$ computed at layer $L$ of DNNs with recurrent architectures
\begin{equation}
P^L(\configvec{f}) = \int \rmd \configvec{h}\,\mathcal{N}(\configvec{h}\vert\configvec{0}, \configvec{c}^{L}  )\prod_{\gamma=1}^{2^n}\delta\big[f_{\gamma},\alpha^{L}(h_{\gamma})\big], \label{eq:P_HL_SL}
\end{equation}
where in the above the element of the covariance matrix is $\big[\configvec{c}^{L}\big]_{\gamma\gamma'} = \big[ \configvec{C} \big]^{L,L}_{\gamma\gamma'} =\sigma^{2} q^{L-1,L-1}_{\gamma\gamma'}$. Note that the \emph{physical meaning} of $P^L(\configvec{f})$ is the distribution of Boolean functions  defined in Eq.~(\ref{def:Boolean-F-density}) averaged over disorder $P^L(\configvec{f})=\lim_{N \to \infty} \overline{P^{L}_{N}(\configvec{f})}$.

Moreover, Eq.~(\ref{eq:P_HL_SL}) also applies to layer-dependent  architectures since the \emph{equal-layer} covariance matrix $\configvec{c}^{L}$ is the same in two scenarios. Therefore we arrive at the first important conclusion that \emph{the typical sets of Boolean functions computed at the output layer $L$ by the layer-dependent  and recurrent architectures are identical}. Furthermore, if the gate functions $\alpha^{l}$ are odd, then it can be shown that all the cross-layer overlaps $q_{\gamma\gamma'}^{l,l'}$ of the recurrent architectures vanish, implying the statistical equivalence of the hidden layer activities to the layered architectures as well~\cite{Mozeika2020sup}.

A similar GF analysis can be applied to sparsely connected Boolean circuits constructed from a single Boolean gate $\alpha$, keeping in mind that distributions of gates can be easily accommodated.  In such models, the source of disorder are random connections. In layer-dependent  architectures, a gate is connected randomly to exactly $k \in O(N^{0})$ gates from the previous layer and this connectivity pattern is changing from layer to layer. In recurrent architectures, on the other hand, the random connections are sampled once and the connectivity pattern is shared among layers. 
Note that in Boolean circuits, the activities at every layer $\configvec{S}_{i}^{l}$ \emph{always} represent a Boolean function.
For layer-dependent  architectures, investigating the distribution of activities gives rise to
\begin{eqnarray}
P^{l+1}(\configvec{f})
&=& \sum_{ \configvec{f}_1,\ldots, \configvec{f}_k}\bigg\{ \prod_{j=1}^{k}P^{l}(\configvec{f}_j)\bigg\}\label{eq:P_S_sparse_layered} \\
&& \times \prod_{\gamma=1}^{2^{n}} \delta\big[f_{\gamma}, \alpha(f_{1,\gamma},\ldots,f_{k,\gamma})\big], \nonumber
\end{eqnarray}
which describes how the probability of the Boolean function $\configvec{f}\in\{-1,1\}^{2^n}$ is evolving from layer to layer~\footnote{Viewing the layers as time steps, the functions can be seen as molecules of gas undergoing $k$-body collisions.}~\cite{Mozeika2020sup}. We note that for recurrent architecture the equation for the probability of Boolean functions computed is exactly the same as above~\cite{Mozeika2020sup}, suggesting that in random Boolean circuits \emph{the  typical sets of Boolean functions computed on layers in the layer-dependent and recurrent architectures are identical}.
Note that the coupling asymmetry plays a crucial role in this equivalence property~\cite{Cessac1995, Hatchett2004, Mozeika2020sup}.

\begin{figure}
    \centering
    \includegraphics[scale=0.25]{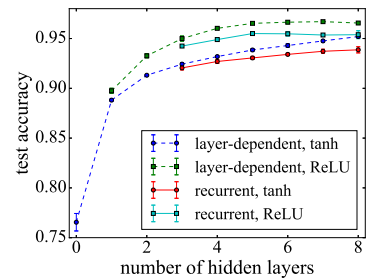}
    \caption{
    Test accuracy of trained fully connected DNNs applied on the MNIST dataset. Images have been downsampled by a factor of $2$ to reduce training time, and each hidden layer has $128$ nodes. Each data point is averaged over $5$ random initializations. The accuracies of  recurrent architectures, with weight Fsharing \emph{between} hidden layers, are comparable to those of layer-dependent  architectures.}
    \label{fig:MNIST_test_acc}
\end{figure}

The equivalence between two architectures points to a potential reduction in the number of free parameters in layered machines by weight sharing or connectivity sharing among layers, useful in devices with limited computation resources~\cite{Cheng2018}. For illustration, we consider the image recognition task of Modified National Institute of Standards and Technology (MNIST) handwritten digit data~\cite{LeCun1998} using DNNs with both layer-dependent and recurrent architectures (weight shared from hidden to hidden layers only; for details see~\cite{Mozeika2020sup}). The experiment shown in Fig.~\ref{fig:MNIST_test_acc} demonstrates the feasibility of using recurrent architectures to perform image classification tasks with a slightly lower accuracy but significant saving in the number of trained parameters.

\textit{Boolean functions computed at large depth.}\textendash\textendash
We consider the typical Boolean functions computed in random-layered machines by examining $P^{L}(\configvec{f})$ in the large depth limit $L\rightarrow\infty$ for specific gates in the following examples.  

In DNNs using the ReLU activation function $\alpha^{l}(x)=\max(x, 0)$, in the hidden layers (the sign activation function is always used in the output layer), which is commonly used in applications, all covariance matrix elements $\left[\configvec{c}^{L}\right]_{\gamma\gamma'}$ in the Eq.~(\ref{eq:P_HL_SL}) converge to the same value in the limit $L \to \infty$, implying that all components of the preactivation field vector $\configvec{h}$ are also the same and hence the components of $\configvec{f}$ are identical. Therefore, random deep ReLU networks compute only \emph{constant} Boolean functions in the infinite depth limit, echoing recent findings of a bias toward simple functions in random DNNs constructed from ReLUs~\cite{Mozeika2020sup}, which arguably plays a role in their generalization ability~\cite{Valle-Perez2018, Palma2019}.

In DNNs using sign activation function also in hidden layers, i.e., Eq.~(\ref{eq:prop}) enforces the rule $S_{i}^{l}=\mathrm{sgn}(\sum_{j} W_{ij}^{l} S_{j}^{l-1}/\sqrt{N})$, those cross-pattern overlaps $q_{\gamma\gamma'}^{l}=(1/N)\sum_{i}\overline{\langle S_{i,\gamma}^{l}S_{i,\gamma'}^{l}\rangle}$ satisfying $|q_{\gamma\gamma'}^{l}|<1$ monotonically decrease with an increasing number of layers and vanish as $l\rightarrow\infty$, such ``chaotic'' nature of dynamics also holds in random DNNs with other sigmoidal activation functions such as the error and hyperbolic tangent functions~\cite{Poole2016, Yang2019}. The consequences of this behavior is that for the input vector  $\sitevec{S}^{I} = \sitevec{s}$, $P^{L}(\configvec
{f})$ is uniform on the set of \emph{all odd} functions~\cite{Mozeika2020sup}, i.e., functions satisfying  $f(-\sitevec{s})=-f(\sitevec{s})$.  Furthermore, for $\sitevec{S}^{I} = (\sitevec{s}, 1)$, $P^{L}(\configvec{f})$ is uniform on the set of \emph{all} Boolean functions~\cite{Mozeika2020sup}.

For Boolean circuits, there are also scenarios where the distribution $P^{L}(\configvec{f})$ has a single Boolean function in its support or it is uniform over some set of functions~\cite{Savicky1990, Brodsky2005, Mozeika2010}. The latter depends on the gates $\alpha$ used in  Eq.~(\ref{eq:prop}) and input vector $\sitevec{S}^{I}$.  For example, in the AND gate with $\alpha(S_{1}, S_{2}) = \mathrm{sgn}(S_{1} + S_{2} +1)$ or the OR gate with $\alpha(S_{1}, S_{2}) = \mathrm{sgn}(S_{1} + S_{2} - 1)$ \cite{Mozeika2020sup}, their output is biased, respectively, toward $+1$ or $-1$~\cite{Savicky1990, Mozeika2010, Mozeika2020sup}. The consequence of the latter is that the distribution $P^{L}(\configvec{f})$ has only a \emph{single} Boolean function in its support~\cite{Mozeika2010, Mozeika2020sup}. On the other hand, when the majority gate $\alpha(S_{1},...,S_{k}) =  \mathrm{sgn}(\sum_{j=1}^{k} S_{j})$, which is balanced $\sum_{S_{1},...,S_{k}} \alpha(S_{1},...,S_{k}) = 0$  and nonlinear~\footnote{Since $\{ S_{j} \}$ are binary variables in the context of Boolean circuits, linearity is defined in the finite field $GF(2)$ \cite{Savicky1990, Brodsky2005}.},  is used with the input vector $\sitevec{S}^{I} = (\sitevec{s}, -\sitevec{s}, 1, -1)$, then the distribution $P^{L}(\configvec{f})$ is uniform over \emph{all Boolean functions}~\cite{Mozeika2010}, which is consistent with the result of~\cite{Savicky1990}.

\textit{Entropy of Boolean functions.}\textendash\textendash
Having considered the distribution of Boolean functions for a few different examples, we observed that random-layered machines either reduce to a single Boolean function or compute all (or a subset of) functions with a uniform probability on the layer $L$, as $L\rightarrow\infty$.  We note that for the Shannon entropy over Boolean functions $\mathcal{H}^{L} = -\sum_{\configvec{f}} P^{L}(\configvec{f}) \log P^{L}(\configvec{f})$, these two scenarios saturate its lower and upper bounds, respectively, given by $0$ and $2^n\log 2$.  Thus, the entropy  $\mathcal{H}^{L}$  can be seen, at least intuitively, as a measure of function space complexity.

\begin{figure}
\includegraphics[scale=0.22]{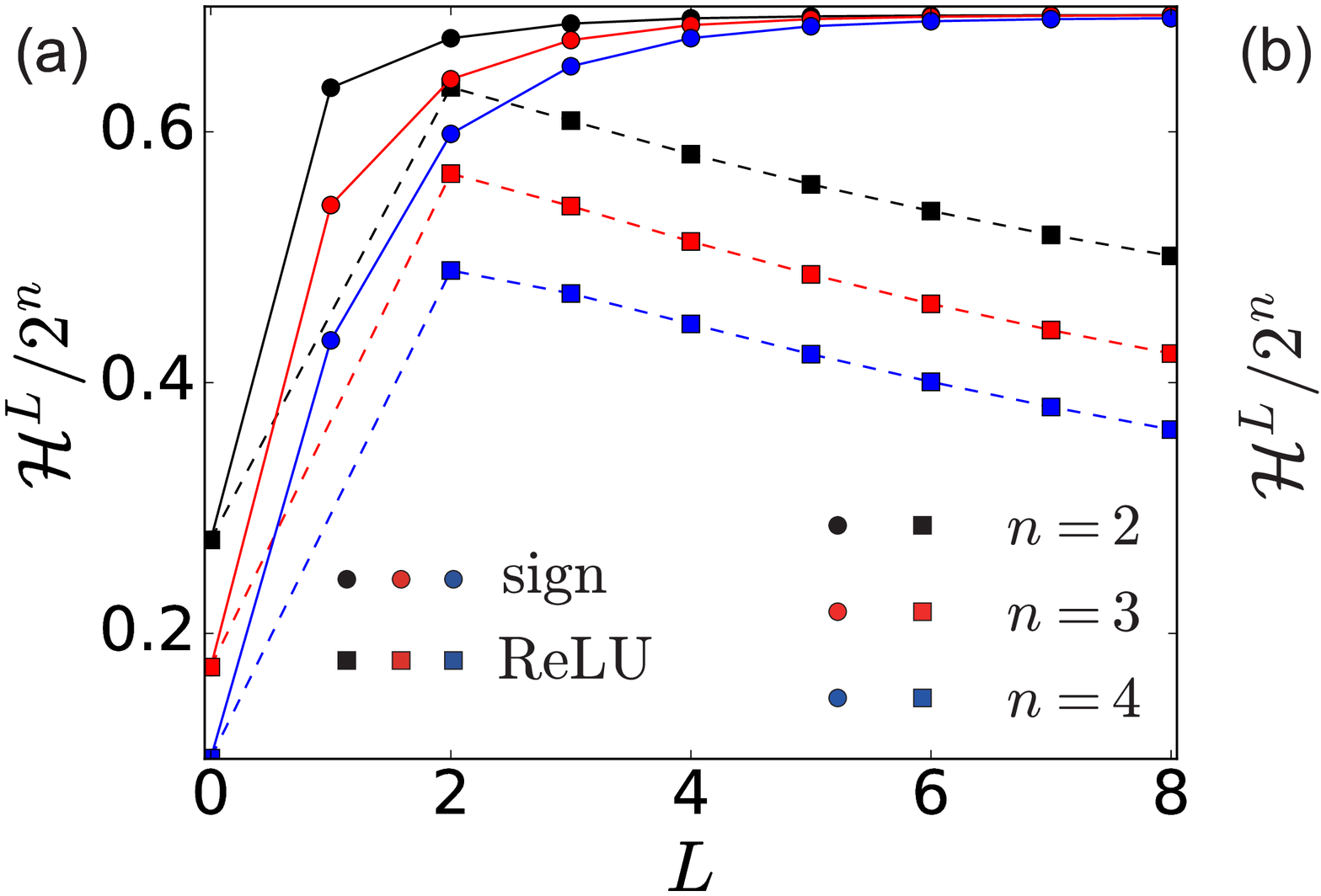}
\includegraphics[scale=0.46]{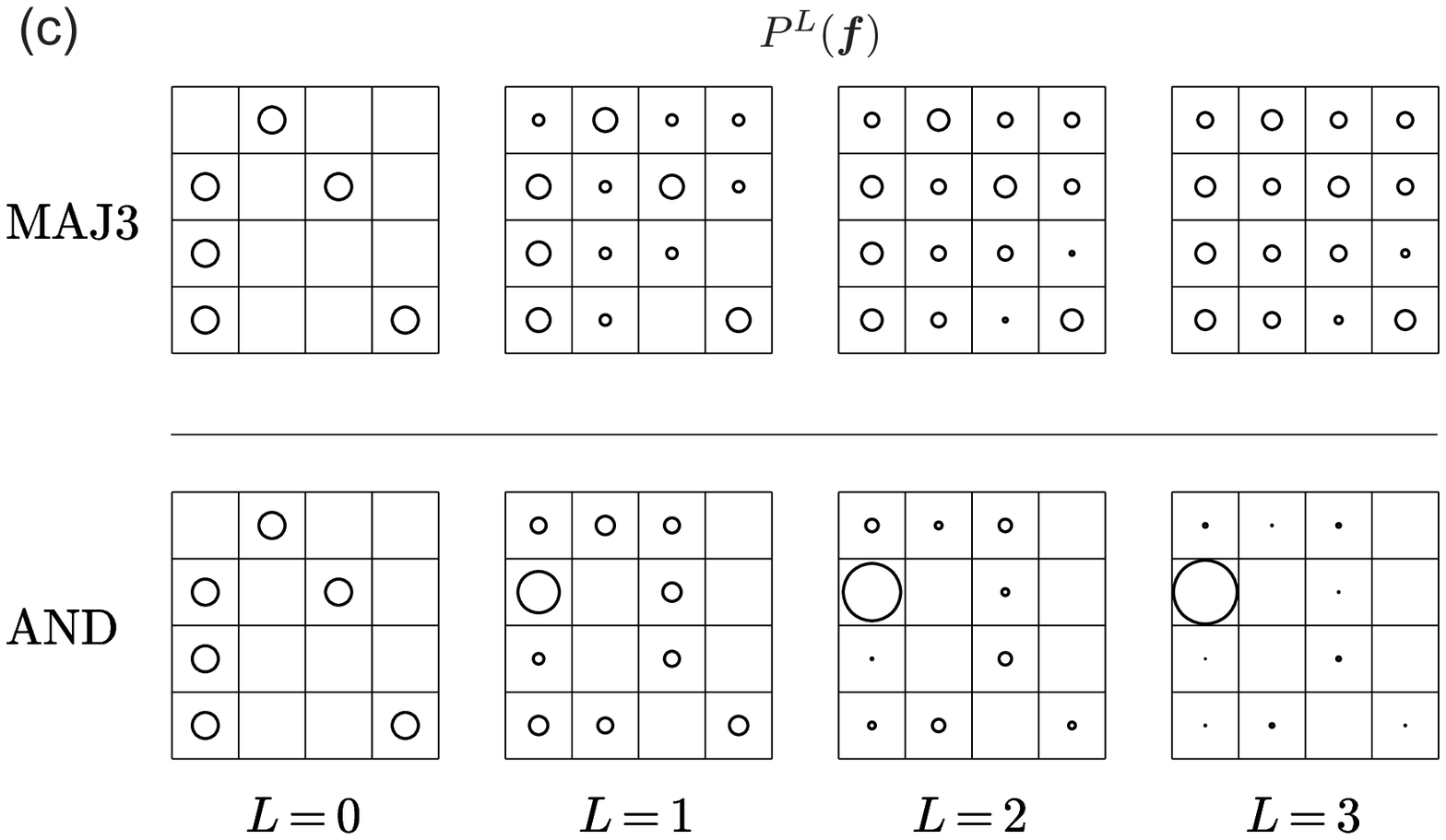}

\caption{Normalized entropy and distribution of functions of deep-layered machines. (a) Normalized entropy $\mathcal{H}^{L}/2^{n}$ of Boolean functions computed by DNNs with sign or ReLU activation in the hidden layers as a function of the network depth $L$; the initial condition is set as $\sitevec{S}^{I} = (\sitevec{s}, 1)$. (b) $\mathcal{H}^{L}/2^{n}$ vs $L$ for Boolean circuits constructed by MAJ3 or the AND gate with initial condition $\sitevec{S}^{I} = (\sitevec{s}, -\sitevec{s}, 1, -1)$. (c) The distribution of Boolean functions $P^{L}(\configvec{f})$ computed by Boolean circuits with two inputs $n=2$ (the number of all possible functions is $16$) is represented by the sizes of circles on a $4\times 4$ grid. Upper panel: MAJ3-gate-based circuits, in which more functions are created at larger depth $L$ and $P^{L}(\configvec{f})$ converges to a uniform distribution. Lower panel: AND-gate-based circuits, in which new functions are created from $L=0$ to $L=1$, while $P^{L}(\configvec{f})$ converges to a distribution with supports in a single Boolean function as network depth increases. \label{fig:ent_S} \vspace*{-5mm}}

\end{figure}

In Fig.~\ref{fig:ent_S}, we study the entropy $\mathcal{H}^{L}$, computed using Eqs.~(\ref{eq:P_HL_SL}) and (\ref{eq:P_S_sparse_layered}), as a function of the depth $L$ in random-layered machines constructed from different activation functions or gates and computing different inputs. The initial increase in entropy after layer $L=0$, seen in Fig.~\ref{fig:ent_S}(a), (b), can be explained by the properties of gates used and the initial set of (simple) Boolean functions at layer $L=0$; functions from the layer $L=0$ are ``copied'' onto layer $L=1$, while new functions are also created, as illustrated in Fig.~\ref{fig:ent_S}(c), (d). Note that the minimal depth in ReLU networks to produce a Boolean function is $L=2$. The dependence of entropy $\mathcal{H}^{L}$ on $L$ after the initial increase depends on the specific gate functions used. For the ReLU activation function in DNNs and the AND gate in Boolean circuits, the entropies $\mathcal{H}^{L}$ monotonically decrease with $L$, suggesting that sizes of sets of typical Boolean functions computed are decreasing with increasing numbers of layers $L$. Random initialization of layered machines with such gates/activation functions serves as a biasing prior toward a more restricted set of functions~\cite{Valle-Perez2018, Palma2019}. On the other hand, for balanced gates, with appropriate initial conditions, e.g., sign in DNNs and majority vote in Boolean circuits, the entropy $\mathcal{H}^{L}$ is monotonically increasing with the depth $L$, indicating that the sizes of sets of the typical Boolean functions computed are increasing.

In summary, we present an analytical framework to examine Boolean functions represented by random deep-layered machines, by considering \emph{all possible inputs simultaneously} and applying the generating functional analysis to compute various relevant macroscopic quantities. We derived the probability of Boolean functions computed on the output nodes. Surprisingly, we discover that the typical sets of Boolean functions computed by the layer-dependent  and recurrent architectures are identical. It points to the possibility of computing complex functions with a reduced number of parameters by weight or connection sharing, as showcased in an image classification experiment. We also study the Boolean functions computed by specific random-layered machines. Biased activation functions (e.g., ReLU) or biased Boolean gates (e.g., AND/OR) can lead to more restricted typical sets of Boolean functions found at deeper layers, which may explain their generalization ability. On the other hand, balanced activation functions (e.g., sign) or Boolean gates (e.g., majority) complemented with appropriate initial conditions, lead to a uniform distribution on all Boolean functions at the infinite depth limit.
It will be interesting to investigate the functions realized by different DNN architectures with structured data and by different learning algorithms~\cite{Saad1998, Goldt2020, Pastore2020, Rotondo2020, Zdeborova2020}.

We also showed the monotonic behavior of the entropy of Boolean functions as a function of depth, which is of interest in the field of computer science. We envisage that the insights gained and the methods developed will facilitate further study of deep-layered machines.

\begin{acknowledgments}
B.L. and D.S. acknowledge support from the Leverhulme Trust (RPG-2018-092), European Union's Horizon 2020 research and innovation program under the Marie Sk{\l}odowska-Curie grant agreement No. 835913. D.S. acknowledges support from the EPSRC program grant TRANSNET (EP/R035342/1). 
\end{acknowledgments}

\bibliographystyle{unsrt}
\bibliography{reference}

\end{document}


\title{Space of Functions Computed by Deep-Layered Machines \\ Supplemental Material}

\author{Alexander Mozeika}
\affiliation{London Institute for Mathematical Sciences, London, W1K 2XF, United Kingdom}

\author{Bo Li}
\affiliation{Nonlinearity and Complexity Research Group, Aston University, Birmingham, B4 7ET, United Kingdom}

\author{David Saad}
\affiliation{Nonlinearity and Complexity Research Group, Aston University, Birmingham, B4 7ET, United Kingdom}

\maketitle

\section{Convention of Notation}

We denote variables with overarrows as vectors with site indices (e.g., $i,j$), which can be of size $k$, $n$ or $N$. On the other hand, we denote bold-symbol variables as vectors of size $2^{n}$ with pattern indices (e.g., $\gamma,\gamma'$), or matrices of size $2^{n} \times 2^{n}$. For convenience, we define $M:=2^{n}$.

The function $\delta(\cdot,\cdot)$ stands for Kronecker delta as $\delta(i,j)=\delta_{i,j}$ if arguments $i,j$ are integer variables, while it stands for Dirac delta function as $\delta(x,y)=\delta(x-y)$ if the arguments $x,y$ are continuous variables; in the latter case, the summation operation should be interpreted as integration, such that $\sum_{y}\delta(x,y)f(y):=\int\mathrm{d}y~\delta(x-y)f(y)$.

The binary variables $S\in\{+1, -1\}$ in this work are mapped onto the conventional Boolean variables $z\in\{0, 1\}$ ($z=0$ represents False, $z=1$ represents True) through $S = 1 - 2z$ ($S=+1$ represents False, $S=-1$ represents True). This choice of notation has the advantage that Boolean addition ($0+0=0, \;\; 0+1=1, \;\; 1+1=0$) can be represented as integer multiplication ($1\times 1=1, \;\; 1\times (-1)=-1, \;\; (-1)\times (-1)=1$). Under this convention, the AND gate is defined as $\mathrm{sgn}(S_i+S_j+1)$, the OR gate is defined as $\mathrm{sgn}(S_i+S_j-1)$, while the majority vote gate is $\mathrm{sgn}(\sum_{j=1}^{k} S_j)$.

\section{Generating Functional Analysis of Fully Connected Neural Networks}

To probe the functions being computed by neural networks, we need
to consider the layer propagation of all $2^{n}$ input patterns
as $\prod_{\gamma=1}^{2^{n}}P(\sitevec{S}_{\gamma}^{L}|\sitevec{S}_{\gamma}^{I}(\sitevec{s}_{\gamma}))$.
We introduce the disorder-averaged generating functional in order
to compute the macroscopic quantities\textbf{ 
\begin{align}
\overline{\Gamma[\{\psi_{i,\gamma}^{l}\}]}= & \overline{ \sum_{\{S_{i,\gamma}^{l}\}_{\forall l,i,\gamma}}\prod_{\gamma=1}^{2^{n}}P(\sitevec{S}_{\gamma}^{0}|\sitevec{S}_{\gamma}^{I})\prod_{l=1}^{L}P(\sitevec{S}_{\gamma}^{l}|\sitevec{S}_{\gamma}^{l-1})e^{-\mathrm{i}\sum_{l,i,\gamma}\psi_{i,\gamma}^{l}S_{i,\gamma}^{l}} }\nonumber \\
= & \mathbb{E}_{W}\sum_{\{S_{i,\gamma}^{l}\}_{\forall l,i,\gamma}}\int\prod_{l=1}^{L}\prod_{i,\gamma}\frac{\mathrm{d}h_{i,\gamma}^{l}\mathrm{d}x_{i,\gamma}^{l}}{2\pi}\prod_{\gamma=1}^{2^{n}}P(\sitevec{S}_{\gamma}^{0}|\sitevec{S}_{\gamma}^{I})\prod_{l=1}^{L}P(\sitevec{S}_{\gamma}^{l}|\sitevec{h}_{\gamma}^{l})e^{-\mathrm{i}\sum_{l,i,\gamma}\psi_{i,\gamma}^{l}S_{i,\gamma}^{l}}\nonumber \\
 & \times\exp\bigg[\sum_{l,i,\gamma}\mathrm{i}x_{i,\gamma}^{l}h_{i,\gamma}^{l}-\sum_{l,\gamma}\sum_{ij}\frac{\mathrm{i}}{\sqrt{N}}W_{ij}^{l}x_{i,\gamma}^{l}S_{j,\gamma}^{l-1}\bigg],\label{eq:gf_def}
\end{align}
}where we have introduced the notation $P(\sitevec{S}_{\gamma}^{l}|\sitevec{h}_{\gamma}^{l})=\prod_{i=1}^{N}P(S_{i,\gamma}^{l}|h_{i,\gamma}^{l})=\prod_{i=1}^{N}\delta\big(S_{i,\gamma}^{l},\alpha^{l}(h_{i,\gamma}^{l})\big)$ and inserted the Fourier representation of unity  $1=\int \frac{\rmd h^{l}_{i,\gamma} \rmd x^{l}_{i,\gamma}}{2 \pi} \exp\big[ \rmi x^{l}_{i,\gamma} \big( h^{l}_{i,\gamma} - \sum_{j} W^{l}_{ij} S^{l-1}_{j,\gamma} \big) \big], \forall l,i,\gamma$.
Noisy computation can be easily accommodated in such probabilistic
formalism.

\subsection{Layer-dependent Architectures}

We first consider layer-dependent weights, where each element follows
the Gaussian distribution $W_{ij}^{l}\sim\mathcal{N}(0,\sigma_{w}^{2})$.
Assuming self-averaging, averaging over the weight disorder component
in the last line of the Eq.~(\ref{eq:gf_def}) yields 
\begin{align}
 & \mathbb{E}_{W}\exp\bigg[-\sum_{l=1}^{L}\sum_{\gamma}\sum_{ij}\frac{\mathrm{i}}{\sqrt{N}}W_{ij}^{l}x_{i,\gamma}^{l}S_{j,\gamma}^{l-1}\bigg]\nonumber \\
= & \exp\bigg[-\frac{\sigma_{w}^{2}}{2}\sum_{l=1}^{L}\sum_{\gamma,\gamma'}\sum_{i}x_{i,\gamma}^{l}x_{i,\gamma'}^{l}\bigg(\frac{1}{N}\sum_{j}S_{j,\gamma}^{l-1}S_{j,\gamma'}^{l-1}\bigg)\bigg].
\end{align}

By introducing the overlap order parameters $\{q_{\gamma\gamma'}^{l}\}_{l=0}^{L}$
through the Fourier representation of unity 
\begin{equation}
1=\int\frac{\mathrm{d}Q_{\gamma\gamma'}^{l}\mathrm{d}q_{\gamma\gamma'}^{l}}{2\pi/N}\exp\bigg[\mathrm{i}NQ_{\gamma\gamma'}^{l}\bigg(q_{\gamma\gamma'}^{l}-\frac{1}{N}\sum_{j}S_{j,\gamma}^{l}S_{j,\gamma'}^{l}\bigg)\bigg],
\end{equation}
the generating functional can be factorized over sites as follows

\textbf{ 
\begin{align*}
\overline{\Gamma[\{\psi_{i,\gamma}^{l}\}]}= & \int\prod_{l=0}^{L}\prod_{\gamma\gamma'}\frac{\mathrm{d}Q_{\gamma\gamma'}^{l}\mathrm{d}q_{\gamma\gamma'}^{l}}{2\pi/N}\exp\bigg[\mathrm{i}N\sum_{l,\gamma\gamma'}Q_{\gamma\gamma'}^{l}q_{\gamma\gamma'}^{l}\bigg]\\
 & \times\exp\bigg[\sum_{i=1}^{N}\log\int\prod_{l=1}^{L}\prod_{\gamma}\mathrm{d}h_{i,\gamma}^{l}\sum_{\{S_{i,\gamma}^{l}\}_{\forall l,\gamma}}\mathcal{M}_{n_{i}}(\configvec{h}_{i},\configvec{S}_{i})\bigg]\\
= & \int\prod_{l=0}^{L}\prod_{\gamma\gamma'}\frac{\mathrm{d}Q_{\gamma\gamma'}^{l}\mathrm{d}q_{\gamma\gamma'}^{l}}{2\pi/N}\exp\bigg[\mathrm{i}N\sum_{l,\gamma\gamma'}Q_{\gamma\gamma'}^{l}q_{\gamma\gamma'}^{l}\bigg]\\
 & \times\exp\bigg[N\bigg(\sum_{m=1}^{|\sitevec{S}^{I}|}\frac{1}{N}\sum_{i=1}^{N}\delta(m,n_{i})\bigg)\log\int\prod_{l=1}^{L}\prod_{\gamma}\mathrm{d}h_{i,\gamma}^{l}\sum_{\{S_{i,\gamma}^{l}\}_{\forall l,\gamma}}\mathcal{M}_{m}(\configvec{h}_{i},\configvec{S}_{i})\bigg],
\end{align*}
}where $\configvec{h}_{i}, \configvec{S}_{i}$ are shorthand notations of $\{ \configvec{h}_{i}^{l} \}, \{ \configvec{S}_{i}^{l} \}$ with $\configvec{h}_{i}^{l} := (h^{l}_{i,1}, ..., h^{l}_{i,\gamma}, ..., h^{l}_{i,2^{n}})$ and $\configvec{S}_{i}^{l} := (S^{l}_{i,1}, ..., S^{l}_{i,\gamma}, ..., S^{l}_{i,2^{n}})$. The single-site measure $\mathcal{M}_{m}$ in the above expression is defined as 
\begin{align}
\mathcal{M}_{m}(\configvec{h}_{i},\configvec{S}_{i})= & \prod_{\gamma=1}^{2^{n}}e^{-\mathrm{i}\sum_{l,\gamma}\psi_{i,\gamma}^{l}S_{i,\gamma}^{l}}P(S_{i,\gamma}^{0}|S_{m,\gamma}^{I})\prod_{l=1}^{L}P(S_{i,\gamma}^{l}|h_{i,\gamma}^{l})\exp\bigg[-\sum_{l,\gamma\gamma'}\mathrm{i}Q_{\gamma\gamma'}^{l}S_{i,\gamma}^{l}S_{i,\gamma'}^{l}\bigg]\nonumber \\
 & \times\prod_{l=1}^{L}\frac{1}{\sqrt{(2\pi)^{2^{n}}|\configvec{c}^{l}|}}\exp\bigg[-\frac{1}{2}\sum_{\gamma\gamma'}h_{i,\gamma}^{l}(\configvec{c}^{l})_{\gamma\gamma'}^{-1}h_{i,\gamma'}^{l}\bigg].\label{eq:Mn_layered}
\end{align}
In Eq.~(\ref{eq:Mn_layered}), $\configvec{c}^{l}$
is a $2^{n}\times 2^{n}$ covariance matrix with elements $\configvec{c}^{l}_{\gamma\gamma'}=\sigma_{w}^{2}q_{\gamma\gamma'}^{l-1}$ and $m$ is a random index following
the empirical distribution $\frac{1}{N}\sum_{i=1}^{N}\delta(m,n_{i})$.

Setting $\psi_{i,\gamma}^{l}=0$ and considering $\lim_{N\to\infty}\frac{1}{N}\sum_{i=1}^{N}\delta(m,n_{i})\to P(m)=1/|\sitevec{S}^{I}|$,
we arrive at

\begin{align}
\overline{\Gamma}= & \int\{\mathrm{d}\configvec{Q}\mathrm{d}\configvec{q}\}e^{N\Psi(\configvec{Q},\configvec{q})},\\
\Psi(\configvec{Q},\configvec{q})= & \sum_{l=0}^{L}\sum_{\gamma\gamma'}\mathrm{i}Q_{\gamma\gamma'}^{l}q_{\gamma\gamma'}^{l}+\sum_{n=1}^{|\sitevec{S}^{I}|}P(n)\log\int\prod_{l=1}^{L}\prod_{\gamma}\mathrm{d}h_{\gamma}^{l}\sum_{\{S_{\gamma}^{l}\}_{\forall l,\gamma}}\mathcal{M}_{m}(\configvec{h},\configvec{S}).
\end{align}
The saddle point equations are obtained by computing $\partial\Psi/\partial q_{\gamma\gamma'}^{l}=0$
and $\partial\Psi/\partial Q_{\gamma\gamma'}^{l}=0$ 
\begin{align}
\mathrm{i}Q_{\gamma\gamma'}^{l-1} & =-\sum_{n}P(n)\frac{\int\mathrm{d}\configvec{h}\sum_{\configvec{S}}\frac{\partial}{\partial q_{\gamma\gamma'}^{l-1}}\mathcal{M}_{m}(\configvec{h},\configvec{S})}{\int\mathrm{d}\configvec{h}\sum_{\configvec{S}}\mathcal{M}_{m}(\configvec{h},\configvec{S})},\quad1\leq l\leq L,\\
\mathrm{i}Q_{\gamma\gamma'}^{L} & =0,\\
q_{\gamma\gamma'}^{l} & =\sum_{m=1}^{|\sitevec{S}^{I}|}P(m)\big\langle S_{\gamma}^{l}S_{\gamma'}^{l}\big\rangle_{\mathcal{M}_{m}},\quad0\leq l\leq L.
\end{align}
Back-propagating the boundary condition $\mathrm{i}Q_{\gamma\gamma'}^{L}=0$
results in $\mathrm{i}Q_{\gamma\gamma'}^{l}=0,\forall l$ \cite{BoLi2018}.

The measure $\mathcal{M}_{m}$ becomes
\begin{align}
\mathcal{M}_{m}(\configvec{h},\configvec{S})= & \prod_{\gamma=1}^{2^{n}}P(S_{\gamma}^{0}|S_{m,\gamma}^{I})\prod_{l=1}^{L}P(S_{\gamma}^{l}|h_{\gamma}^{l})\nonumber \\
 & \times\prod_{l=1}^{L}\frac{1}{\sqrt{(2\pi)^{2^{n}}|\configvec{c}^{l}|}}\exp\bigg[-\frac{1}{2}\sum_{\gamma\gamma'}h_{\gamma}^{l}(\configvec{c}^{l})_{\gamma\gamma'}^{-1}h_{\gamma'}^{l}\bigg],
\end{align}
while the saddle point equations of overlaps have the form of

\begin{align}
q_{\gamma\gamma'}^{0} & =\sum_{m}P(m)S_{m,\gamma}^{I}S_{m,\gamma'}^{I},\label{eq:q0_layered}\\
q_{\gamma\gamma'}^{l} & =\int\mathrm{d}h_{\gamma}^{l}\mathrm{d}h_{\gamma'}^{l}\frac{\phi^{l}(h_{\gamma}^{l})\phi^{l}(h_{\gamma'}^{l})}{\sqrt{(2\pi)^{2}|\Sigma_{\gamma\gamma'}^{l}|}}\exp\bigg[-\frac{1}{2}[h_{\gamma}^{l},h_{\gamma'}^{l}]\cdot(\Sigma_{\gamma\gamma'}^{l})^{-1}\cdot[h_{\gamma}^{l},h_{\gamma'}^{l}]^{\top}\bigg],\label{eq:ql_layered}
\end{align}
where the $2\times2$ covariance matrix $\Sigma_{\gamma\gamma'}^{l}$ is defined as 
\begin{align}
\Sigma_{\gamma\gamma'}^{l} & :=\sigma_{w}^{2}\left(\begin{array}{cc}
q_{\gamma\gamma}^{l-1} & q_{\gamma\gamma'}^{l-1}\\
q_{\gamma'\gamma}^{l-1} & q_{\gamma'\gamma'}^{l-1}
\end{array}\right).
\end{align}

\subsection{Recurrent Architectures}

In this section, we consider recurrent topology where the weights
are independent of layers $W_{ij}^{l}=W_{ij}\sim\mathcal{N}(0,\sigma_{w}^{2})$.
The calculation resembles the case of layer-dependent weights, except that
the disorder average yields cross-layer overlaps

\begin{align}
 & \mathbb{E}_{W}\exp\bigg[-\sum_{l=1}^{L}\sum_{\gamma}\sum_{ij}\frac{\mathrm{i}}{\sqrt{N}}W_{ij}x_{i,\gamma}^{l}S_{j,\gamma}^{l-1}\bigg]\nonumber \\
= & \exp\bigg[-\frac{\sigma_{w}^{2}}{2}\sum_{l,l'=1}^{L}\sum_{\gamma,\gamma'}\sum_{i}x_{i,\gamma}^{l}x_{i,\gamma'}^{l'}\bigg(\frac{1}{N}\sum_{j}S_{j,\gamma}^{l-1}S_{j,\gamma'}^{l'-1}\bigg)\bigg].
\end{align}
Introducing order parameters $q_{\gamma\gamma'}^{l,l'}:=\frac{1}{N}\sum_{j}S_{j,\gamma}^{l}S_{j,\gamma'}^{l'}$
and setting $\psi_{i,\gamma}^{l}=0$, we eventually obtain
\begin{align}
\overline{\Gamma}= & \int\{\mathrm{d}\configvec{Q}\mathrm{d}\configvec{q}\}e^{N\Psi(\configvec{Q},\configvec{q})},\\
\Psi(\configvec{Q},\configvec{q})= & \rmi\Tr\left\{\configvec{q}\,\configvec{Q} \right\} +\sum_{m=1}^{|\sitevec{S}^{I}|}P(m)\log\int\prod_{l=1}^{L}\prod_{\gamma}\mathrm{d}h_{\gamma}^{l}\sum_{\{S_{\gamma}^{l}\}_{\forall l,\gamma}}\mathcal{M}_{m}(\configvec{h},\configvec{S}),\\
\mathcal{M}_{m}(\configvec{h},\configvec{S})= & \prod_{\gamma=1}^{2^{n}}P(S_{\gamma}^{0}|S_{m,\gamma}^{I})\prod_{l=1}^{L}P(S_{\gamma}^{l}|h_{\gamma}^{l})\exp\bigg[-\sum_{ll',\gamma\gamma'}\mathrm{i}Q_{\gamma\gamma'}^{l,l'}S_{\gamma}^{l}S_{\gamma'}^{l'}\bigg]\nonumber \\
 & \times\frac{1}{\sqrt{(2\pi)^{2^{n}L}|\configvec{C}|}}\exp\bigg[-\frac{1}{2}\configvec{H}^{\top}\configvec{C}^{-1}\configvec{H}\bigg],
\end{align}
where $\rmi\Tr\left\{\configvec{q}\,\configvec{Q} \right\} = \rmi \sum_{l,l'=0}^{L}\sum_{\gamma\gamma'} Q_{\gamma\gamma'}^{l,l'}q_{\gamma\gamma'}^{l,l'}$ and $\configvec{H} = (\configvec{h}^{1}, ..., \configvec{h}^{L}) \in \R^{2^{n}L}$ expresses the preactivation fields of all patterns and all layers,
while $\configvec{C}$ is a $2^{n}L\times2^{n}L$ covariance
matrix.

The corresponding saddle point equations are 
\begin{align}
\mathrm{i}Q_{\gamma\gamma'}^{l-1,l'-1} & =-\sum_{n}P(n)\frac{\int\mathrm{d}\configvec{h}\sum_{\configvec{S}}\frac{\partial}{\partial q_{\gamma\gamma'}^{l-1,l'-1}}\mathcal{M}_{m}(\configvec{h},\configvec{S})}{\int\mathrm{d}\configvec{h}\sum_{\configvec{S}}\mathcal{M}_{m}(\configvec{h},\configvec{S})},\quad1\leq l,l'\leq L,\\
\mathrm{i}Q_{\gamma\gamma'}^{L,l} & =0,\quad\forall l\\
q_{\gamma\gamma'}^{l,l'} & =\sum_{m}P(m)\big\langle S_{\gamma}^{l}S_{\gamma'}^{l'}\big\rangle_{\mathcal{M}_{m}},\quad0\leq l\leq L.
\end{align}

All conjugate order parameters $\{\mathrm{i}Q_{\gamma\gamma'}^{l,l'}\}$
vanish identically similar to the previous case, such that the effective
single-site measure becomes 
\begin{align}
\mathcal{M}_{m}(\configvec{h},\configvec{S})= & \prod_{\gamma=1}^{2^{n}}P(S_{\gamma}^{0}|S_{m,\gamma}^{I})\prod_{l=1}^{L}P(S_{\gamma}^{l}|h_{\gamma}^{l})\nonumber \\
 & \times\frac{1}{\sqrt{(2\pi)^{2^{n}L}|\configvec{C}|}}\exp\bigg[-\frac{1}{2}\configvec{H}^{\top}\configvec{C}^{-1}\configvec{H}\bigg],
\end{align}
and the saddle point equation of the order parameters follows 
\begin{align}
q_{\gamma\gamma'}^{0,0} & =\sum_{m}P(m)S_{m,\gamma}^{I}S_{m,\gamma'}^{I},\label{eq:q00_recurrent}\\
q_{\gamma\gamma'}^{l,0} & =\sum_{m}P(m)\big\langle S_{\gamma}^{l}S_{\gamma'}^{0}\big\rangle_{\mathcal{M}_{m}}\nonumber \\
 & =\bigg(\sum_{m}P(m)S_{m,\gamma'}^{I}\bigg)\int\mathrm{d}h_{\gamma}^{l}\frac{\phi^{l}(h_{\gamma}^{l})}{\sqrt{2\pi\sigma_{w}^{2}}}\exp\bigg[-\frac{1}{2\sigma_{w}^{2}}\big(h_{\gamma}^{l}\big)^{2}\bigg],\label{eq:ql0_recurrent}\\
q_{\gamma\gamma'}^{l,l'} & =\int\mathrm{d}h_{\gamma}^{l}\mathrm{d}h_{\gamma'}^{l'}\frac{\phi^{l}(h_{\gamma}^{l})\phi^{l'}(h_{\gamma'}^{l'})}{\sqrt{(2\pi)^{2}|\Sigma_{\gamma\gamma'}^{l,l'}|}}\exp\bigg[-\frac{1}{2}[h_{\gamma}^{l},h_{\gamma'}^{l'}]\cdot(\Sigma_{\gamma\gamma'}^{l,l'})^{-1}\cdot[h_{\gamma}^{l},h_{\gamma'}^{l'}]^{\top}\bigg],\label{eq:qllp_recurrent}
\end{align}
where the $2\times2$ covariance matrix $\Sigma_{\gamma\gamma'}^{l,l'}$
is defined as 
\begin{align}
\Sigma_{\gamma\gamma'}^{l,l'} & :=\sigma_{w}^{2}\left(\begin{array}{cc}
q_{\gamma\gamma}^{l-1,l-1} & q_{\gamma\gamma'}^{l-1,l'-1}\\
q_{\gamma'\gamma}^{l'-1,l-1} & q_{\gamma'\gamma'}^{l'-1,l'-1}
\end{array}\right).
\end{align}
Similar formalism was derived in the context of dynamical recurrent
neural networks to study the autocorrelation of spin/neural dynamics
\cite{Toyoizumi2015}.

\subsection{Strong Equivalence Between Layer-dependent and Recurrent Architectures for
Odd Activation Functions}

In general, the statistical properties of the activities of machines
of layer-dependent architectures and recurrent architectures are different,
since the fields $\{\configvec{h}^{l}\}$ of different layers are
directly correlated in the latter case. However, one can observe that
the equal-layer overlaps $q_{\gamma\gamma'}^{l,l}$ in the recurrent
architectures is identical to $q_{\gamma\gamma'}^{l}$ in the layer-dependent
architectures, by noticing the same initial condition in Eq.~(\ref{eq:q00_recurrent})
and Eq.~(\ref{eq:q0_layered}) and the same forward propagation rules
in Eq.~(\ref{eq:qllp_recurrent}) (with $l'=l$) and Eq.~(\ref{eq:ql_layered}).

If the cross-layer overlaps $\{q_{\gamma\gamma'}^{l,l'}|l\neq l'\}$
vanish, then the direct correlation between $\configvec{h}^{l}$
of different layers also vanish such that 
\begin{equation}
\frac{1}{\sqrt{(2\pi)^{2^{n}L}|\configvec{C}|}}\exp\bigg[-\frac{1}{2}\configvec{H}^{\top}\configvec{C}^{-1}\configvec{H}\bigg]=\prod_{l}\frac{1}{\sqrt{(2\pi)^{2^{n}}|\configvec{c}^{l}|}}\exp\bigg[-\frac{1}{2}(\configvec{h}^{l})^{\top}(\configvec{c}^{l})^{-1}\configvec{h}^{l}\bigg].
\end{equation}
In this case, the distributions of the macroscopic trajectories $\{\configvec{h}^{l},\configvec{S}^{l}\}$
of the two architectures are equivalent. One sufficient
condition for this to hold is that the activation functions $\phi^{l}(\cdot)$
are odd functions satisfying $\phi^{l}(-x)=-\phi^{l}(x)$. Firstly,
this condition implies that $q_{\gamma\gamma'}^{l,0}=0,\forall l$
by Eq.~(\ref{eq:ql0_recurrent}); secondly, $q_{\gamma\gamma'}^{l,0}=0$
and the fact that $\phi^{l}(\cdot)$ is odd implies $q_{\gamma\gamma'}^{l+1,1}=0$,
which leads to $q_{\gamma\gamma'}^{l,l'}=0,\forall l\neq l'$ by induction.

\subsection{Weak Equivalence Between Layer-dependent and Recurrent Architectures for
General Activation Functions}

As shown above, in general the trajectories $\{\configvec{h}^{l},\configvec{S}^{l}\}$
of layer-dependent architectures follow a different
distribution from the case of recurrent architectures with shared
weights except for some specific cases such as DNNs with odd activation
functions. Here we focus on the distribution of activities in the
output layer.

For layer-dependent weights, the joint distribution of the local fields
and activations at layer $L$ is obtained by marginalizing the variables
of initial and hidden layers 
\begin{align}
P(\configvec{h}^{L},\configvec{S}^{L}) & =\int\prod_{\gamma}\frac{\mathrm{d}x_{\gamma}^{L}}{2\pi}\int\prod_{l=1}^{L-1}\prod_{\gamma}\mathrm{d}h_{\gamma}^{l}\sum_{m}P(m)\sum_{\{S_{\gamma}^{l}\}_{\forall\gamma,l<L}}\mathcal{M}_{m}(\configvec{h},\configvec{S})\nonumber \\
 & =\int\prod_{l=1}^{L-1}\mathrm{d}\configvec{h}^{l}\sum_{\{S_{\gamma}^{l}\}_{\forall\gamma,l<L}}\bigg(\prod_{l=1}^{L}\prod_{\gamma}P(S_{\gamma}^{l}|h_{\gamma}^{l})\bigg)\prod_{l=1}^{L}\frac{1}{\sqrt{(2\pi)^{2^{n}}|\configvec{c}^{l}|}}\exp\bigg[-\frac{1}{2}(\configvec{h}^{l})^{\top}(\configvec{c}^{l})^{-1}\configvec{h}^{l}\bigg]\nonumber \\
 & =\mathcal{N}(\configvec{h}^{L}|\configvec{0},\configvec{c}^{L}(\configvec{q}^{L-1}))\prod^{2^{n}}_{\gamma=1}P(S_{\gamma}^{L}|h_{\gamma}^{L}),
\end{align}
where $\mathcal{N}(\configvec{h}^{L}|\configvec{0},\configvec{c}^{L}(\configvec{q}^{L-1}))$ is a $2^{n}$ dimensional multivariate Gaussian distribution. The distribution of Boolean functions $f(\cdot)$ computed at layer $L$ is
\begin{eqnarray}
    P^L(\configvec{f}) &=& \int \rmd \configvec{h}^{L} \sum_{\configvec{S}^{L}} P(\configvec{h}^{L},\configvec{S}^{L}) \prod_{\gamma=1}^{2^{n}} \delta(S_{\gamma}^{L}, f(\sitevec{s}_{\gamma})) \nonumber \\
    &=& \int \rmd \configvec{h} \mathcal{N}(\configvec{h}|\configvec{0},\configvec{c}^{L}) \prod_{\gamma=1}^{2^{n}} \delta(f_{\gamma}, \alpha^{L}(h_{\gamma})),
\end{eqnarray}
where the binary string $\configvec{f}$ of size $2^{n}$ represents the Boolean function $f(\cdot)$ with $f_{\gamma} = f(\sitevec{s}_{\gamma})$.

For shared weights, the fields of all layers $\configvec{H} = (\configvec{h}^{1}, ..., \configvec{h}^{l}, ..., \configvec{h}^{L}) \in \R^{2^{n}L}$
are coupled with covariance $\configvec{C}$
\begin{align}
P(\configvec{h}^{L},\configvec{S}^{L}) & =\int\prod_{l=1}^{L-1}\mathrm{d}\configvec{h}^{l}\sum_{\{S_{\gamma}^{l}\}_{\forall\gamma,l<L}}\bigg(\prod_{l=1}^{L}\prod_{\gamma}P(S_{\gamma}^{l}|h_{\gamma}^{l})\bigg)\frac{1}{\sqrt{(2\pi)^{2^{n}L}|\configvec{C}|}}\exp\bigg[-\frac{1}{2}(\configvec{H})^{\top}\configvec{C}^{-1}\configvec{H}\bigg]\nonumber \\
 & =\prod_{\gamma}P(S_{\gamma}^{L}|h_{\gamma}^{L})\int\prod_{l=1}^{L-1}\mathrm{d}\configvec{h}^{l}\frac{1}{\sqrt{(2\pi)^{2^{n}L}|\configvec{C}|}}\exp\bigg[-\frac{1}{2}(\configvec{H})^{\top}\configvec{C}^{-1}\configvec{H}\bigg]\nonumber \\
 & =\prod_{\gamma}P(S_{\gamma}^{L}|h_{\gamma}^{L})\frac{1}{\sqrt{(2\pi)^{2^{n}}|\configvec{C}^{L,L}|}}\exp\bigg[-\frac{1}{2}(\configvec{h}^{L})^{\top}(\configvec{C}^{L,L})^{-1}\configvec{h}^{L}\bigg]\nonumber \\
 & =\mathcal{N}(\configvec{h}^{L}|\configvec{0},\configvec{C}^{L,L}(\configvec{q}^{L-1,L-1}))\prod^{2^{n}}_{\gamma=1}P(S_{\gamma}^{L}|h_{\gamma}^{L}).
\end{align}
Since the equal-layer overlap follows the same dynamical rule with
the case of layer-dependent weights such that $\configvec{C}^{L, L} = \configvec{c}^{L}$, the distributions $P(\configvec{h}^{L},\configvec{S}^{L})$ of the two
scenarios are equivalent. This suggests that if only the input-output mapping
is of interest (but not the hidden layer activity), the distributions of the Boolean functions $P^{L}(\configvec{f})$ computed at the final layer of the two architectures are equivalent.

\section{Remark on the Equivalence Property}
We observed that although auto-correlations generally exist in recurrent architectures, they do not complicate the single-layer macroscopic behaviors of the system studied. This is due to the fact that the weights/couplings being used are asymmetric (i.e., $W_{ij}$ and $W_{ji}$ are independent of each other), such that there are no intricate feedback interactions of a node with its state at previous time steps, which renders a Markovian single-layer macroscopic dynamics~\cite{Cessac1995}. If symmetric couplings are present, then the whole history of the network is needed to characterize the dynamics~\cite{Cessac1995}. Similar effect of coupling asymmetry was also observed in sparsely connected networks~\cite{Hatchett2004}. Early works investigating asymmetric coupling in neural networks include~\cite{Derrida1987, Kree1987}. 

Although we only consider finite input dimension, we expect that the equivalence property also holds in the cases where $n$ has the same order as $N$ as long as the couplings are asymmetric.

\section{Extension of the Theory on Densely-connected Neural Networks}
While the theory on densely-connected neural networks was developed in the infinite width limit $N \to \infty$, we expect that it applies to large but finite systems (some properties investigated below require $N \gg n$). In~\cite{BoLi2020}, it is found that the order parameters in such systems satisfy the large deviation principle, which implies an exponential convergence rate to the typical behaviors as the width $N$ grows. Typically, order parameters of systems with $N \sim 10^3$ are well concentrated around their typical values as predicted by the theory.

Accommodating the cases where $n$ has the same order as $N$ (both tend to infinity) in the current framework is a bit subtle, as it requires an infinite number of order parameters and may result in the loss of the self-averaging properties.

While Boolean input variables are of primary interest here, the input domain can be generalized to any countable sets; this is relevant for adapting the theory to real input variables with defined numerical precision, e.g., if a real input variable $s \in [0, 1]$ is subject to precision of $0.01$, then only a finite number of possible values of $s \in \{0, 0.01, 0.02, ..., 1 \}$ need to be considered. On the other hand, real input variables with arbitrary precision are difficult to deal with, as there is an uncountably infinite number of input patterns and the product $\prod_{\gamma} \cdots$ is ill-defined. It is worthwhile noting~\cite{Yang2019} that random neural networks on real spherical data shares similar properties with those on Boolean data in high dimension; therefore we expect that the equivalence between layer-dependent and recurrent architectures also applies to real data.

Other than being able to treat the highly correlated recurrent architectures, the GF or path-integral framework also facilitates the characterization of fluctuations around the typical behaviors~\cite{Crisanti2018}, and the computation of large deviations in finite-size systems~\cite{BoLi2020}.

\section{Training Experiments}
\begin{figure}
    \centering
    \includegraphics[scale=0.32]{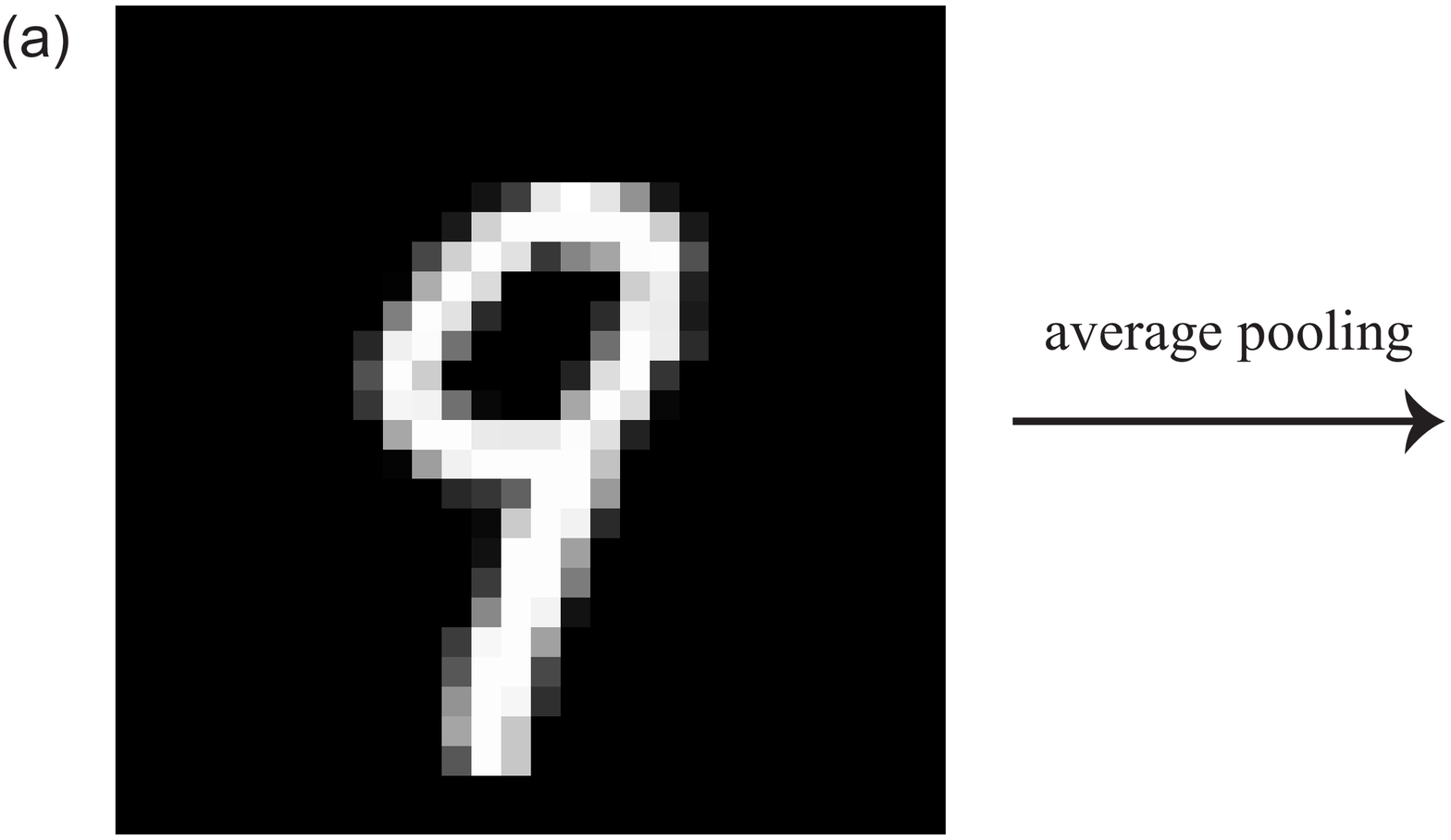}
    \caption{(a) The MNIST data are pre-processed by average pooling to downsample the images, in order to reduce training time. (b) Test accuracy of trained fully connected DNNs with 6 hidden layers applied to MNIST dataset. For different widths of the hidden layers, DNNs with recurrent architectures can achieve comparable performance to those with layer-dependent architectures. }
    \label{fig:MNIST}
\end{figure}

Figure 2 of the main text demonstrates the feasibility of using DNNs with recurrent architectures to perform an image recognition task on the MNIST hand-written digit data. In this section, we describe the details of the training experiment. The objective is not to achieve state-of-the-art performance, but to showcase the potential in using recurrent architectures for parameter reduction. Therefore, we pre-process the image data by downsampling with a factor of 2 through average pooling, which saves runtime of training by reducing the size of each image from $28\times 28$ to $14\times 14$. See Fig.~\ref{fig:MNIST}(a) for an example.

We consider DNNs of both architectures, layer-dependent and recurrent, where the input $\sitevec{s}$ is directly copied onto the initial layer $\sitevec{S}^{0}$, and a softmax function is applies to the final layer. We remark that the theory developed in this work is applicable to random-weight DNNs implementing Boolean functions, while it is not directly applicable to trained networks.  For recurrent architectures, since the dimension of input and output layers are fixed, only weights $W^{\text{hid}}$ between hidden layers are shared, i.e.
\begin{equation}
    \sitevec{S}^{0}\xrightarrow{W^{\text{in}}}\sitevec{S}^{1}\xrightarrow{W^{\text{hid}}}\sitevec{S}^{2}\xrightarrow{W^{\text{hid}}}\cdots \sitevec{S}^{l}\xrightarrow{W^{\text{hid}}}\sitevec{S}^{l+1}\xrightarrow{W^{\text{hid}}}\cdots\xrightarrow{W^{\text{hid}}}\sitevec{S}^{L-1}\xrightarrow{W^{\text{out}}}\sitevec{S}^{L},
\end{equation}
where all the hidden layers have the same width. The corresponding DNNs of both architectures are trained by the ADAM algorithm with back-propagation~\cite{Kingma2015}. In Fig.~\ref{fig:MNIST}(b), we demonstrate that for different widths of hidden layers, DNNs with recurrent architectures can achieve performance that is comparable to those with layer-dependent architectures.

\section{Boolean Functions Computed by Random DNNs}

To examine the distribution of Boolean functions computed at
layer $L$ (we always apply sign activation function in
the final layer), notice that nodes at layer $L$ are not coupled
together, so it is sufficient to consider a particular node in the
final layer, which follows the distribution of the effective single
site measure established before.

Further notice that the local field $\configvec{h}^{L} \in \R^{2^{n}}$
in the final layer follows a multivariate Gaussian distribution with
zero mean and covariance $\configvec{c}_{\gamma\gamma'}^{L}=\sigma_{w}^{2}q_{\gamma\gamma'}^{L-1}$.
Essentially the local field $h^{L}$ is a Gaussian process with a
dot product kernel (in the limit $N\to\infty$) \cite{Lee2018,Yang2019}
\begin{equation}
k(\sitevec{x},\sitevec{x}')=k\bigg(\frac{\sitevec{x}\cdot\sitevec{x}'}{n}\bigg)=\sigma_{w}^{2}q_{x,x'}^{L-1},
\end{equation}
where $\text{\ensuremath{\sitevec{x}},\ensuremath{\sitevec{x}'}}$
are $n$-dimensional vectors.

The probability of a Boolean function $f(s_{1,\gamma},...,s_{n,\gamma})$
being computed in the fully connected neural network is

\begin{align}
P^{L}(\configvec{f})  =\int\mathrm{d}\configvec{h}\mathcal{N}(\configvec{h}|\configvec{0},\configvec{c}^{L}(\configvec{q}))\prod_{\gamma=1}^{2^{n}}\delta\big(\mathrm{sgn}(h_{\gamma}^{L}), f_{\gamma}\big).
\end{align}
We focus on systems of layer-dependent architectures, where the overlap $q_{\gamma\gamma'}^{l}$
is governed by the forward dynamics

\begin{equation}
q_{\gamma\gamma'}^{l}=\int\mathrm{d}h_{\gamma}^{l}\mathrm{d}h_{\gamma'}^{l}\frac{\phi(h_{\gamma}^{l})\phi(h_{\gamma'}^{l})}{\sqrt{(2\pi)^{2}|\Sigma_{l}|}}\exp\bigg[-\frac{1}{2}[h_{\gamma}^{l},h_{\gamma'}^{l}]\cdot\big(\Sigma_{\gamma\gamma'}^{l}(q^{l-1})\big)^{-1}\cdot[h_{\gamma}^{l},h_{\gamma'}^{l}]^{\top}\bigg].
\end{equation}

\begin{itemize}
\item For sign activation function, choosing $\sigma_{w}=1$ yields 
\begin{equation}
\Sigma_{\gamma\gamma'}^{l}=\left(\begin{array}{cc}
1 & q_{\gamma\gamma'}^{l-1}\\
q_{\gamma\gamma'}^{l-1} & 1
\end{array}\right),
\end{equation}
\end{itemize}
\begin{align}
q_{\gamma\gamma'}^{l} & =\frac{2}{\pi}\sin^{-1}\big(q_{\gamma\gamma'}^{l-1}\big),\quad\forall l>0,\\
q_{\gamma\gamma'}^{0} & =\begin{cases}
\frac{1}{n}\sum_{m=1}^{n}s_{m,\gamma}s_{m,\gamma'}, & \sitevec{S}^{I}=\sitevec{s},\\
\frac{1}{n+1}\sum_{m=1}^{n}(1+s_{m,\gamma}s_{m,\gamma'}). & \sitevec{S}^{I}=(\sitevec{s},1).
\end{cases}
\end{align}

\begin{figure}
\includegraphics[scale=0.4]{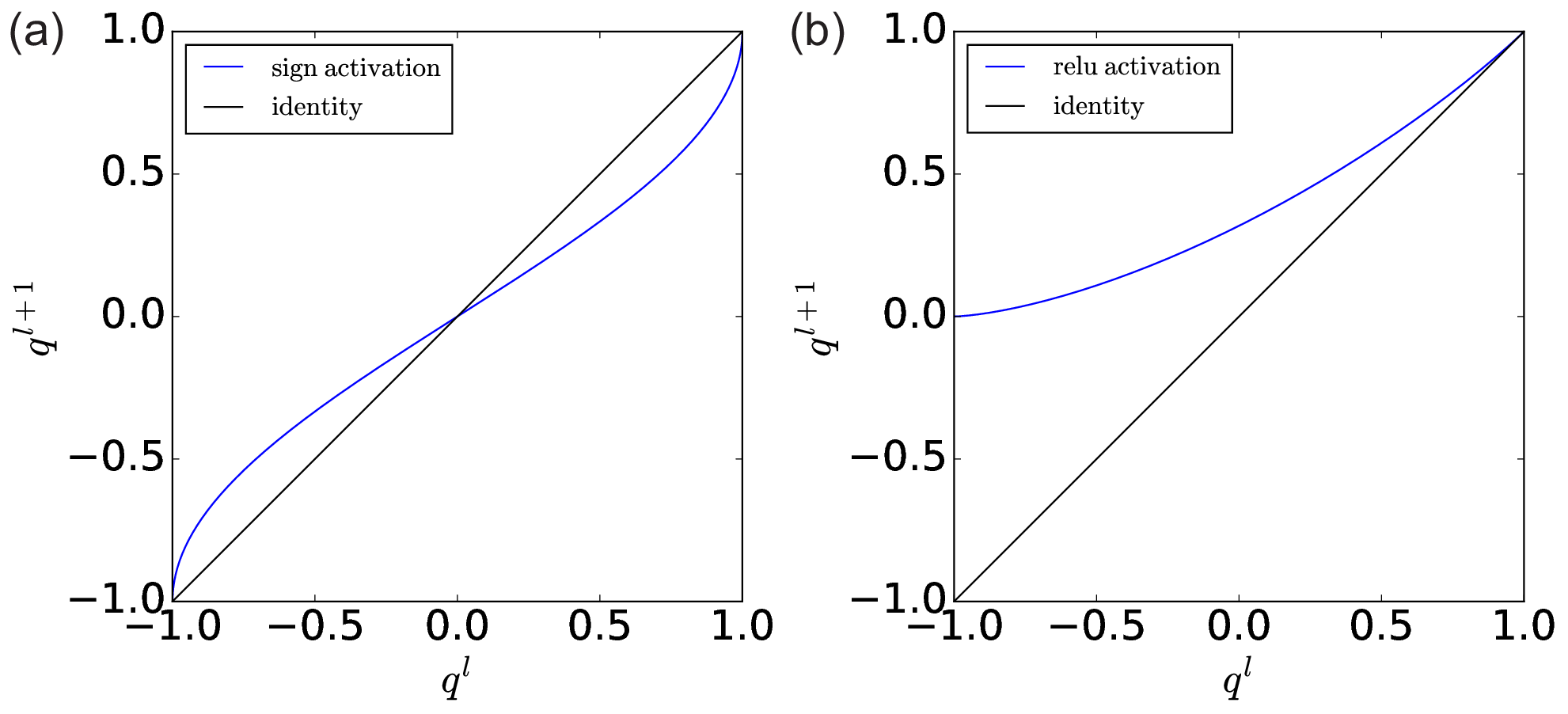}

\caption{Iteration mappings of overlaps of fully connected neural networks in the absense of bias variables. (a) sign activation function. $q^{l}=0$
is a stable fixed point while $q^{l}=1,-1$ are two unstable fixed
points. (b) ReLU activation functions with $\sigma_{w}=\sqrt{2}$.
$q^{l}=1$ is a stable fixed point.\label{fig:iter_map}}
\end{figure}

\begin{itemize}
\item For ReLU activation function, choosing $\sigma_{w}=\sqrt{2}$ yields
\begin{equation}
\Sigma_{\gamma\gamma'}^{l}=2\left(\begin{array}{cc}
q_{\gamma\gamma}^{l-1} & q_{\gamma\gamma'}^{l-1}\\
q_{\gamma\gamma'}^{l-1} & q_{\gamma'\gamma'}^{l-1}
\end{array}\right)=2\left(\begin{array}{cc}
q_{\gamma\gamma}^{l-1} & q_{\gamma\gamma'}^{l-1}\\
q_{\gamma\gamma'}^{l-1} & q_{\gamma'\gamma'}^{l-1}
\end{array}\right),
\end{equation}
\begin{align}
q_{\gamma\gamma}^{l} & =\frac{1}{2}\Sigma_{\gamma\gamma}^{l}=1,\quad\forall l,\gamma\\
q_{\gamma\gamma'}^{l} & =\frac{1}{2\pi}\bigg[\sqrt{\big|\Sigma_{\gamma\gamma'}^{l}\big|}+\frac{\pi}{2}\Sigma_{\gamma\gamma',12}^{l}+\Sigma_{\gamma\gamma',12}^{l}\tan^{-1}\bigg(\Sigma_{\gamma\gamma',12}^{l}\big/\sqrt{\big|\Sigma_{\gamma\gamma'}^{l}\big|}\bigg)\bigg]\nonumber \\
 & =\frac{1}{\pi}\bigg[\sqrt{1-\big(q_{\gamma\gamma'}^{l-1}\big)^{2}}+q_{\gamma\gamma'}^{l-1}\bigg(\frac{\pi}{2}+\sin^{-1}\big(q_{\gamma\gamma'}^{l-1}\big)\bigg)\bigg],\forall l>0,\\
q_{\gamma\gamma'}^{0} & =\begin{cases}
\frac{1}{n}\sum_{m=1}^{n}s_{m,\gamma}s_{m,\gamma'}, & \sitevec{S}^{I}=\sitevec{s},\\
\frac{1}{n+1}\sum_{m=1}^{n}(1+s_{m,\gamma}s_{m,\gamma'}). & \sitevec{S}^{I}=(\sitevec{s},1).
\end{cases}
\end{align}
\end{itemize}
The iteration mappings of overlaps of the two activation functions
considered are depicted in Fig.~\ref{fig:iter_map}.

\subsection{ReLU Networks in the Large $L$ Limit}

In this section, we focus on the large depth limit $L\to\infty$.
For ReLU activation function, all the matrix elements of $\configvec{c}^{L}$
become identical in the large $L$ limit, leading to $\configvec{c}^{L}(\configvec{q})\propto J$
(where $J$ is the all-one matrix) and a degenerate Gaussian distribution
of the vector $\configvec{h}^{L}$ enforcing all its components to be
the same. To make it explicit, we consider the distribution of $\configvec{h}^{L}$
as follows
\begin{align}
P(\configvec{h}^{L}) & =\frac{1}{\sqrt{(2\pi)^{M}|\configvec{c}^{L}|}}\exp\bigg[-\frac{1}{2}(\configvec{h}^{L})^{\top}(\configvec{c}^{L})^{-1}\configvec{h}^{L}\bigg],\nonumber \\
 & =\int\frac{\mathrm{d}\configvec{x}^{L}}{(2\pi)^{M}}\exp\bigg(\mathrm{i}\configvec{x}^{L}\cdot\configvec{h}^{L}-\frac{\sigma_{w}^{2}}{2}(\configvec{x}^{L})^{\top}J\configvec{x}^{L}\bigg)\\
 & =\lim_{\kappa\to1}\int\frac{\mathrm{d}\configvec{x}^{L}}{(2\pi)^{M}}\exp\bigg(\mathrm{i}\configvec{x}^{L}\cdot\configvec{h}^{L}-\frac{\sigma_{w}^{2}}{2}\big[\sum_{\gamma}(x_{\gamma}^{L})^{2}+\kappa\sum_{\gamma\neq\gamma'}x_{\gamma}^{L}x_{\gamma'}^{L}\big]\bigg),
\end{align}
Now define $\configvec{c}(\kappa)=\sigma_{w}^{2}\big[(1-\kappa)I+\kappa J\big]$
and notice

\begin{equation}
[\configvec{c}(\kappa)]^{-1}=\frac{1}{\sigma_{w}^{2}(1-\kappa)}\bigg(I-\frac{\kappa}{M\rho+(1-\kappa)}J\bigg)\approx\frac{1}{\sigma_{w}^{2}(1-\kappa)}\bigg(I-\frac{1}{M}J+\frac{1-\kappa}{M^{2}\kappa}J\bigg),
\end{equation}
\begin{align}
P(\configvec{h}^{L}) & =\lim_{\kappa\to1}\frac{1}{\sqrt{(2\pi)^{M}|\configvec{c}(\kappa)|}}\exp\bigg(-\frac{1}{2}(\configvec{h}^{L})^{\top}[\configvec{c}(\kappa)]^{-1}\configvec{h}^{L}\bigg)\nonumber \\
 & =\lim_{\kappa\to1}\frac{1}{\sqrt{(2\pi)^{M}|\configvec{c}(\kappa)|}}\exp\bigg\{-\frac{1}{2}\frac{M}{\sigma_{w}^{2}(1-\kappa)}\bigg(\frac{1}{M}\sum_{\gamma}(h_{\gamma}^{L})^{2}-\big(\frac{1}{M}\sum_{\gamma}h_{\gamma}^{L}\big)^{2}\bigg)\bigg\}\nonumber \\
 & \qquad\quad\times\exp\bigg\{-\frac{1}{2\sigma_{w}^{2}\kappa}\big(\frac{1}{M}\sum_{\gamma}h_{\gamma}^{L}\big)^{2}\bigg\}.
\end{align}
In the limit $\kappa\to1$, $P(\configvec{h}^{L})$ has support only
in the subspace with $\frac{1}{M}\sum_{\gamma}(h_{\gamma}^{L})^{2}-\big(\frac{1}{M}\sum_{\gamma}h_{\gamma}^{L}\big)^{2}=0$,
requiring all $h_{\gamma}^{L}$ to be identical.

Therefore, the output node computes a constant Boolean function in
the limit $L\to\infty$ 
\begin{equation}
P^{L}(\configvec{f})=\frac{1}{2},\quad\text{with }f(\cdot)=1\text{ or }f(\cdot)=-1.
\end{equation}

\subsection{Sign Networks in the Large $L$ Limit}

\subsubsection{Input Vector $\sitevec{S}^{I}=(s_{1},s_{2},...,s_{n})$}

For the input vector $\sitevec{S}^{I}=(s_{1},s_{2},...,s_{n})$, the
overlap at layer $0$ is computed as

\begin{equation}
q_{\gamma\gamma'}^{0}=\frac{1}{n}\sum_{m=1}^{n}s_{m,\gamma}s_{m,\gamma'},
\end{equation}
which satisfies $-1\leq q_{\gamma\gamma'}^{0}\leq1$; $q_{\gamma\gamma'}^{0}=1$
iff $\gamma=\gamma'$, while $q_{\gamma\gamma'}^{0}=-1$ iff $\text{\ensuremath{\sitevec{s}}}_{\gamma'}=-\text{\ensuremath{\sitevec{s}}}_{\gamma}$
(input $\gamma'$ is the negation of input $\gamma$).

We label the $M=2^{n}$ patterns according to 
\begin{align}
\gamma & =1+\sum_{m=1}^{n}\frac{1-s_{m,\gamma}}{2}2^{n-m},
\end{align}
\begin{equation}
s_{m,\gamma}=1-2\times\text{mod}\bigg( \Bigl\lfloor \frac{\gamma-1}{2^{n-m}} \Bigr\rfloor,2\bigg).
\end{equation}
We note that, the mapping from $\gamma$ to $s_{m,\gamma}$
is as follows: (i) represent the integer $\gamma-1\in\{0,1,...,M-1\}$
by its binary string; for example, for $n=3$, the $M=8$ configurations
are arranged in the ordered $[000,001,010,011,100,101,110,111]$;
(ii) turn the binary variable $0(1)$ at each site of the binary
string into Ising variable $+1(-1)$ or $+(-)$.

Under this convention, for two negating inputs $\gamma,\gamma'$ $\text{\ensuremath{\sitevec{s}}}_{\gamma'}=-\text{\ensuremath{\sitevec{s}}}_{\gamma}$,
the indices satisfied

\begin{align}
\gamma+\gamma' & =2+\sum_{m=1}^{n}\bigg(\frac{1-s_{m,\gamma}}{2}+\frac{1+s_{m,\gamma}}{2}\bigg)2^{n-m}\nonumber \\
 & =2+2^{n}-1=M+1.
\end{align}

In the large $L$ limit, all matrix elements of $\configvec{c}^{L}$
vanish except the diagonal terms $\Sigma_{\gamma\gamma}^{L}$ and
anti-diagonal terms $\Sigma_{\gamma\gamma'}^{L}\delta_{\gamma+\gamma',M+1}$.
For instance, for $n=2$, the covariance matrix $\Sigma^{L}$ has
the following structure 
\begin{equation}
\Sigma^{L}/\sigma_{w}^{2}=\begin{array}{c|cccc}
 & ++ & +- & -+ & --\\
\hline ++ & 1 & 0 & 0 & -1\\
+- & 0 & 1 & -1 & 0\\
-+ & 0 & -1 & 1 & 0\\
-- & -1 & 0 & 0 & 1
\end{array},
\end{equation}
which is singular and corresponds to a degenerate Gaussian distribution
of $\configvec{h}^{L}$. Essentially, it implies that the input $\gamma$ and its negation $\gamma'$ are anti-correlated in the local
fields $\langle h_{\gamma}^{L}h_{\gamma'}^{L}\rangle/\sigma_{w}^{2}=-1$.

To make the constraints of the degenerate Gaussian distribution explicit,
we consider replacing the anti-diagonal elements of $\configvec{c}^{L}$
by $\kappa$, make use of the identities in Sec. \ref{sec:matrix_Am},
and take the limit $\kappa\to-1$ in the end of the calculation,

\begin{align}
P(\configvec{h}^{L}) & =\lim_{\kappa\to-1}\int\frac{\mathrm{d}\configvec{x}^{L}}{(2\pi)^{M}}\exp\bigg(\mathrm{i}\configvec{x}^{L}\cdot\configvec{h}^{L}-\frac{\sigma_{w}^{2}}{2}(\configvec{x}^{L})^{\top}\cdot A_{M}(\kappa)\cdot\configvec{x}^{L}\bigg),\nonumber \\
 & =\lim_{\kappa\to-1}\frac{1}{\sqrt{(2\pi)^{M}\sigma_{w}^{2M}(1-\kappa^{2})^{M/2}}}\exp\bigg(-\frac{1}{2\sigma_{w}^{2}}(\configvec{h}^{L})^{\top}\cdot A_{M}(\kappa)^{-1}\cdot\configvec{h}^{L}\bigg),
\end{align}

\begin{align}
(\configvec{h}^{L})^{\top}A_{M}(\kappa)^{-1}\configvec{h}^{L} & =\frac{1}{1-\kappa^{2}}\sum_{\mu,\nu=1}^{M}h_{\mu}^{L}h_{\nu}^{L}(\delta_{\mu\nu}-\kappa\delta_{\mu+\nu,M+1})\nonumber \\
 & =\frac{1}{1-\kappa^{2}}\sum_{\mu,\nu=1}^{M}\big[\frac{1}{2}(h_{\mu}^{L})^{2}+\frac{1}{2}(h_{\nu}^{L})^{2}-\kappa h_{\mu}^{L}h_{\nu}^{L}\big]\delta_{\mu+\nu,M+1}\nonumber \\
 & =\frac{1}{1-\kappa^{2}}\sum_{\mu,\nu=1}^{M}\big[\frac{1}{2}(h_{\mu}^{L}+h_{\nu}^{L})^{2}-(1+\kappa)h_{\mu}^{L}h_{\nu}^{L}\big]\delta_{\mu+\nu,M+1},
\end{align}
\begin{align}
P(\configvec{h}^{L}) & =\lim_{\kappa\to-1}\frac{1}{\sqrt{(2\pi)^{M/2}\sigma_{w}^{M}(1-\kappa^{2})^{M/2}}}\exp\bigg(-\frac{1}{2\sigma_{w}^{2}(1-\kappa^{2})}\sum_{\gamma=1}^{M/2}(h_{\gamma}^{L}+h_{M+1-\gamma}^{L})^{2}\bigg)\nonumber \\
 & \qquad\quad\times\frac{1}{\sqrt{(2\pi)^{M/2}\sigma_{w}^{M}}}\exp\bigg(-\frac{1}{2\sigma_{w}^{2}(1-\kappa)}2\sum_{\gamma=1}^{M/2}h_{\gamma}^{L}h_{M+1-\gamma}^{L}\bigg)\nonumber \\
 & =\prod_{\gamma=1}^{M/2}\bigg\{\delta\big(h_{\gamma}^{L}+h_{M+1-\gamma}^{L}\big)\frac{1}{\sqrt{2\pi\sigma_{w}^{2}}}\exp\big[-\frac{1}{2\sigma_{w}^{2}}(h_{\gamma}^{L})^{2}\big]\bigg\}.
\end{align}
Therefore, the first $\frac{M}{2}$ fields are independent of each
other, while the last $\frac{M}{2}$ fields have the opposite sign. The probability
of a Boolean function $F$ being computed is
\begin{align}
P^{L}(\configvec{f}) & =\int\mathrm{d}\configvec{h}^{L}\mathcal{N}(\configvec{h}^{L}|0,\Sigma^{L})\prod_{\gamma}\delta\big(\mathrm{sgn}(h_{\gamma}^{L}),f(\sitevec{S}_{\gamma}^{I})\big)\nonumber \\
 & =\prod_{\gamma=1}^{M/2}\int\mathrm{d}h_{\gamma}^{L}\mathcal{N}(h_{\gamma}^{L}|0,\sigma_{w}^{2})\delta\big(\mathrm{sgn}(h_{\gamma}^{L}),f(\sitevec{S}_{\gamma}^{I})\big)\delta\big(\mathrm{sgn}(-h_{\gamma}^{L}),f(\sitevec{S}_{M+1-\gamma}^{I})\big)\nonumber \\
 & =\prod_{\gamma=1}^{M/2}\frac{1}{2}\mathbb{I}\big(f(\sitevec{s}_{M+1-\gamma})=-f(\sitevec{s}_{\gamma})\big)\nonumber \\
 & =\frac{1}{\sqrt{2^{M}}}\prod_{\gamma=1}^{M/2}\mathbb{I}\big(f(-\sitevec{s}_{\gamma})=-f(\sitevec{s}_{\gamma})\big),
\end{align}
where $\mathbb{I}(\cdot)$ is the indicator function returning 1 if the condition is met and zero otherwise. The space of functions computed is uniformly distributed among all the odd functions (negated
inputs lead to negated output). The restriction to odd functions can be understood by the fact that both $h^{l}_{i}(\sitevec{S}^{l-1}) = \sum_{j} W^{l}_{ij} S^{l-1}_{j}$
and $\mathrm{sgn}(h^{l}_{i})$ in the forward propagation are odd functions,
which imposes a symmetry constraint in the input-output mappings.
Such symmetry is broken by the bias in the input vector $\sitevec{S}^{I}=(\sitevec{s},1)$ as is shown in the following section.

\subsubsection{Input Vector $\sitevec{S}^{I}=(s_{1},s_{2},...,s_{n},1)$}

If we consider the input vector $\sitevec{S}^{I}=(\sitevec{s},1)=(s_{1},s_{2},...,s_{n},1)$,
the overlap at layer $0$ is given by 
\begin{align}
q_{\gamma\gamma'}^{0} & =\frac{1}{n+1}\bigg(1+\sum_{m=1}^{n}s_{m,\gamma}s_{m,\gamma'}\bigg),
\end{align}
which satisfies $-1<q_{\gamma\gamma'}^{0}\leq1$ and $q_{\gamma\gamma'}^{0}=1$
iff $\gamma=\gamma'$. This choice of input set is equivalent to adding
a bias variable in the first layer. For the sign activation function,
$q^{l}=0$ is a stable fixed point unless $q^{0}=1$. Therefore, in
the large $L$ limit, all off-diagonal matrix elements of the
$2^{n}$-dimensional covariance matrix vanish, leading to $\configvec{c}^{L}(\configvec{q})\propto I$, the identity matrix. The distribution of functions
computed at the output node is 
\begin{align}
P^{L}(\configvec{f}) & =\int\mathrm{d}\configvec{h}^{L}\mathcal{N}(\configvec{h}^{L}|0,\sigma_{w}^{2}I)\prod_{\gamma}\delta\big(\mathrm{sgn}(h_{\gamma}^{L}),f(\sitevec{S}_{\gamma}^{I})\big)\nonumber \\
 & =\prod_{\gamma}\int\mathrm{d}h_{\gamma}^{L}\mathcal{N}(h_{\gamma}^{L}|0,\sigma_{w}^{2})\delta\big(\mathrm{sgn}(h_{\gamma}^{L}),f(\sitevec{S}_{\gamma}^{I})\big)\nonumber \\
 & =\prod_{\gamma=1}^{2^{n}}\frac{1}{2}=\frac{1}{2^{2^{n}}}
\end{align}
i.e. the uniform distribution over \emph{all} Boolean functions.

In the above analysis, if the input dimension $n$ is large, the smallest overlap $q_{\gamma\gamma'}^{0}$ can get very close to $-1$ and convergence to the uniform distribution of functions (as network depth increase) becomes very slow. In this case, one can set the biased input vector as $\sitevec{S}^{I}=(\sitevec{s},const)=(s_{1},s_{2},...,s_{n}, const)$ where the constant variable satisfies $const \in O(n)$, such that $q_{\gamma\gamma'}^{0}$ deviates from $-1$ and the convergence to a uniform distribution over all Boolean functions becomes faster.

\subsubsection{Introducing Bias Variables}

\begin{figure}
\includegraphics[scale=0.4]{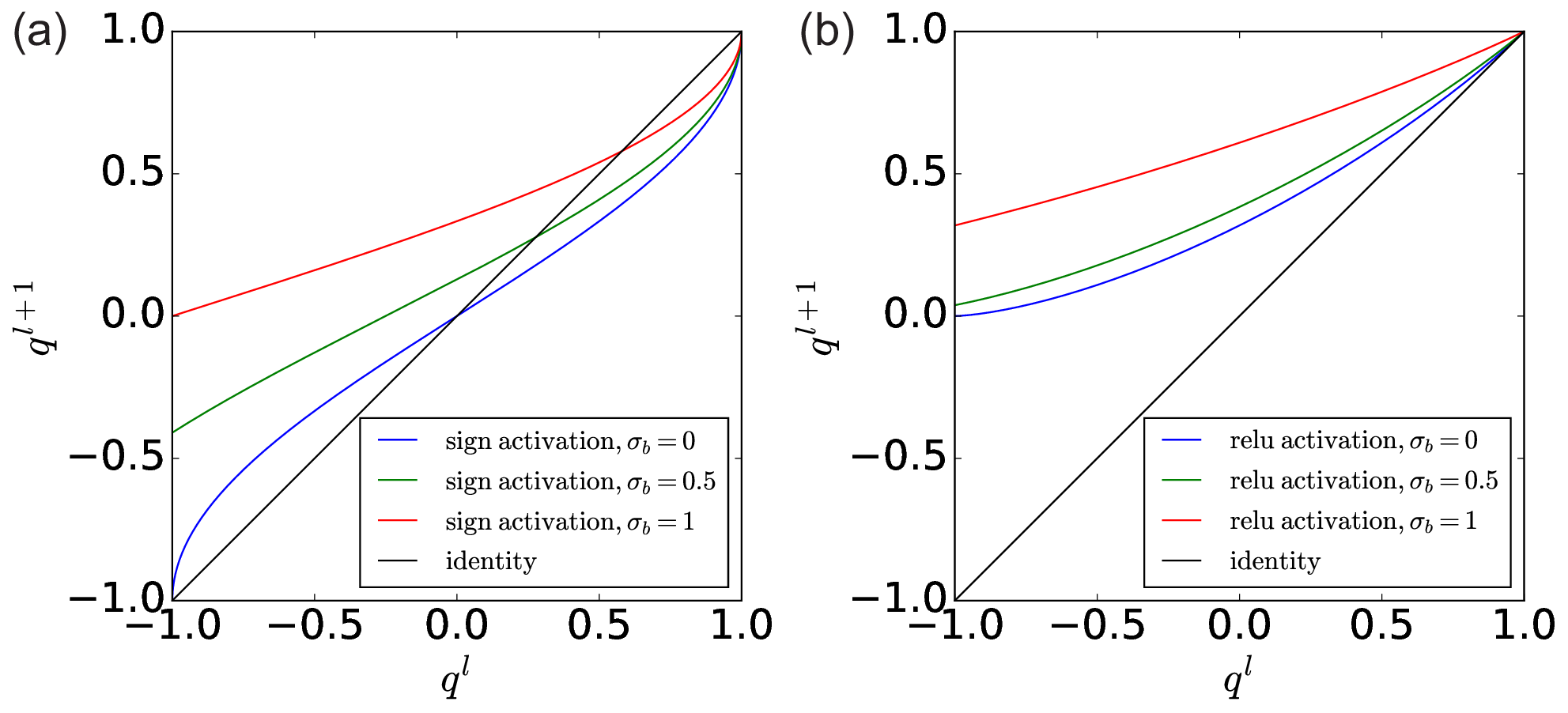}

\caption{Iteration mappings of overlaps of fully connected neural networks in the presence of bias variables. (a) sign activation function with
$\sigma_{w}=1$. The overlap $q^{l}=1$ is an unstable fixed point, and there
exists a stable fixed point with $q^{*}>0$. (b) ReLU activation functions
with $\sigma_{w}^{2}=2-\sigma_{b}^{2}$. $q^{l}=1$ is a stable fixed
point. It shows how the fixed point drifts to a higher value with the increasing bias. \label{fig:iter_map_wb}}
\end{figure}

In this section, we introduce the conventional bias variables of neural
networks as follows
\begin{equation}
h_{i}^{l}(W^{l}\cdot\sitevec{S}^{l-1})=\frac{1}{\sqrt{N}}\sum_{j=1}^{N}W_{ij}^{l}S_{j}^{l-1}+b_{i}^{l},
\end{equation}
where $W_{ij}^{l}\sim\mathcal{N}(0,\sigma_{w}^{2})$ and $b_{i}^{l}\sim\mathcal{N}(0,\sigma_{b}^{2})$.
The only difference from the case without bias is the average

\begin{align}
 & \mathbb{E}_{W,b}\exp\bigg[-\sum_{l=1}^{L}\sum_{\gamma}\sum_{ij}\frac{\mathrm{i}}{\sqrt{N}}W_{ij}^{l}x_{i,\gamma}^{l}S_{j,\gamma}^{l-1}-\sum_{l=1}^{L}\sum_{\gamma}\sum_{i}\mathrm{i}x_{i,\gamma}^{l}b_{i}^{l}\bigg]\nonumber \\
= & \exp\bigg[-\frac{1}{2}\sum_{l=1}^{L}\sum_{\gamma,\gamma'}\sum_{i}x_{i,\gamma}^{l}x_{i,\gamma'}^{l}\bigg(\frac{\sigma_{w}^{2}}{N}\sum_{j}S_{j,\gamma}^{l-1}S_{j,\gamma'}^{l-1}+\sigma_{b}^{2}\bigg)\bigg].
\end{align}
This leads to the following overlap dynamics 
\begin{itemize}
\item Sign activation 
\begin{equation}
q_{\gamma\gamma'}^{l}=\frac{2}{\pi}\sin^{-1}\bigg(\frac{\sigma_{w}^{2}q_{\gamma\gamma'}^{l-1}+\sigma_{b}^{2}}{\sigma_{w}^{2}+\sigma_{b}^{2}}\bigg),\quad\forall l>0,
\end{equation}
\item ReLU activation with $\sigma_{w}^{2}+\sigma_{b}^{2}=2$ (to ensure
$q_{\gamma\gamma}^{l}=1$) 
\begin{equation}
q_{\gamma\gamma'}^{l}=\frac{1}{\pi}\bigg[\sqrt{1-\bigg(\frac{\sigma_{w}^{2}q_{\gamma\gamma'}^{l-1}+\sigma_{b}^{2}}{2}\bigg)^{2}}+\frac{\sigma_{w}^{2}q_{\gamma\gamma'}^{l-1}+\sigma_{b}^{2}}{2}\bigg(\frac{\pi}{2}+\sin^{-1}\bigg(\frac{\sigma_{w}^{2}q_{\gamma\gamma'}^{l-1}+\sigma_{b}^{2}}{2}\bigg)\bigg)\bigg],\forall l>0,
\end{equation}
\end{itemize}
The iteration mappings of overlaps considered are depicted in Fig.
\ref{fig:iter_map_wb}.

\subsection{Finite $L$}

Perhaps the more interesting scenario is the case with finite $L$.
The entropy of the Boolean functions that the output node computes
is 
\begin{align}
\mathcal{H}^{L}= & -\sum_{\configvec{f}}P^{L}(\configvec{f})\log P^{L}(\configvec{f})\nonumber \\
= & -\sum_{\{f_{\gamma}\}_{\forall\gamma}}\int\mathrm{d}\configvec{h}^{L}\mathcal{N}(\configvec{h}^{L}|\configvec{0},\configvec{c}^{L})\prod_{\gamma}\delta\big(f_{\gamma},\mathrm{sgn}(h_{\gamma}^{L})\big)\nonumber \\
 & \qquad\times\log\int\mathrm{d}\configvec{h}^{L}\mathcal{N}(\configvec{h}^{L}|\configvec{0},\configvec{c}^{L})\prod_{\gamma}\delta\big(f_{\gamma},\mathrm{sgn}(h_{\gamma}^{L})\big),\label{eq:def_entropy}
\end{align}
which is fully determined by the covariance matrix $\configvec{c}^{L}(\configvec{q}^{L-1})$.

\begin{figure}
\includegraphics[scale=0.4]{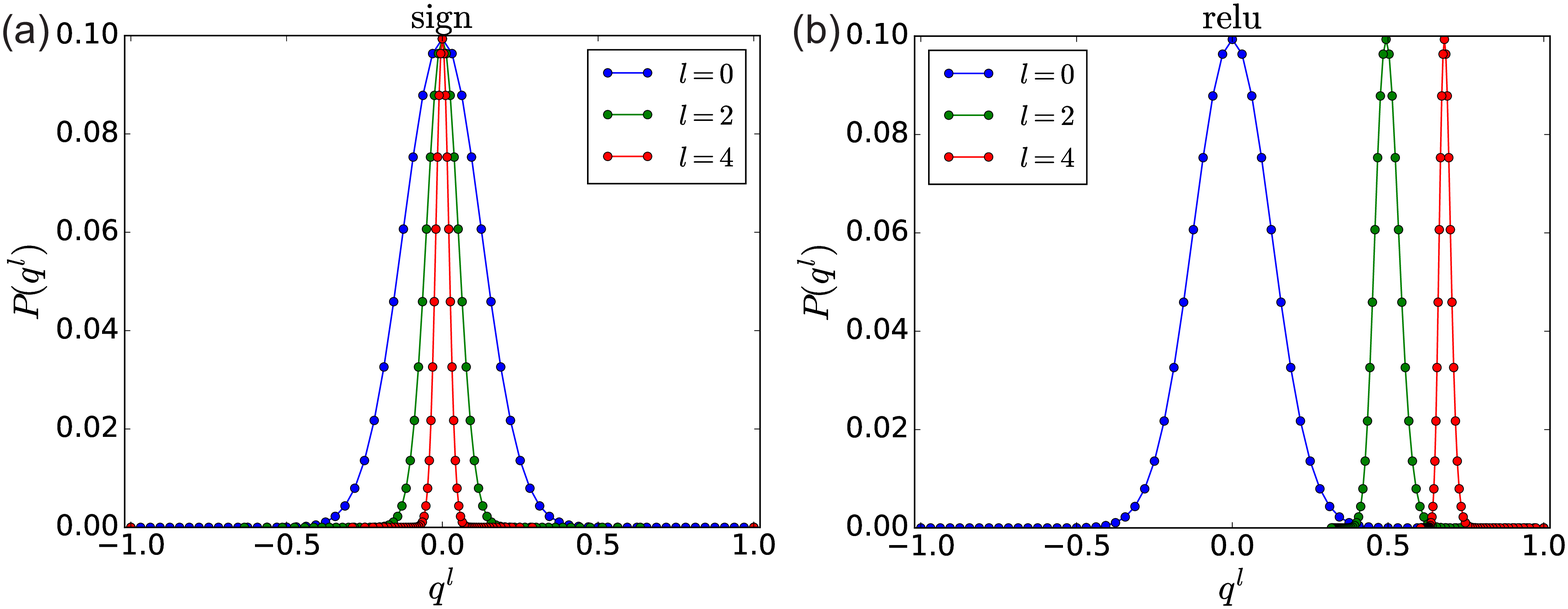}

\caption{Frequency of appearance of matrix elements $q_{\gamma\gamma'}^{l}$
in $\configvec{c}^{l+1}$. $n=64$. (a) Sign activation. (b) ReLU
activation. \label{fig:Pql} }
\end{figure}

For input set of $\sitevec{S}^{I}=(s_{1},s_{2},...,s_{n})$, we
have $q_{\gamma\gamma'}^{0}\in\{-1,-1+\frac{2}{n},...,1\}$; the forward
propagation rule implies that the overlap $q_{\gamma\gamma'}^{l}$
at any layer $l$ has only $n+1$ possible values. The Hamming distance
between $s_{\gamma}^{I}$ and $s_{\gamma'}^{I}$ is $d_{\gamma\gamma'}=\frac{n}{2}(1-q_{\gamma\gamma'}^{0})$,
therefore at each row/column of the matrix $\configvec{c}^{L}$, there
are ${d_{\gamma\gamma'} \choose n}$ elements which take the value
$q_{\gamma\gamma'}^{L-1}$. Fig.~\ref{fig:Pql} depicts the frequency
of $q_{\gamma\gamma'}^{l}$ in different layers defined as

\begin{equation}
P(q^{L})={d_{\gamma\gamma'} \choose n}\bigg/2^{n}.
\end{equation}

\begin{figure}
\includegraphics[scale=0.5]{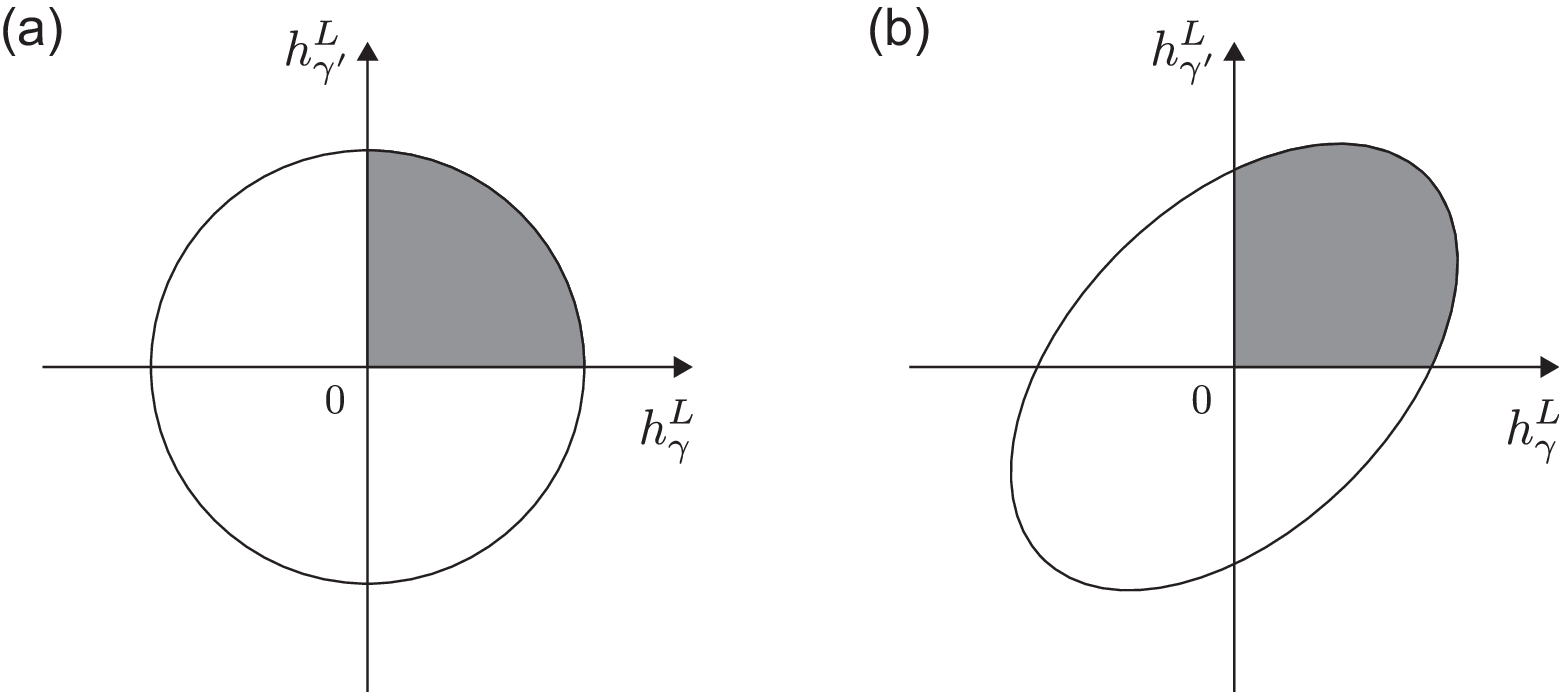}

\caption{Local field distribution in the case of $n=1$. (a) Uncorrelated Gaussian
distribution. The probability mass in the first quadrant corresponds
to the function $f(\cdot)=1$ being computed with probability
$\frac{1}{4}$. (b) Correlated Gaussian distribution. The function
$f(\cdot)=1$ appears with a probability larger than $\frac{1}{4}$.
\label{fig:h_illustrate}}
\end{figure}

Intuitively, the local fields $h_{\gamma}^{L}$ that
are more correlated lead to a lower entropy of
the Boolean functions computed. See Fig.~\ref{fig:h_illustrate} for
an illustration of the case of single variable input. It shows a gradual concentration (in layers) around zero of the overlap for sign activation-based layered networks and a concentration while drifting away toward one of the overlap value in the ReLU case. We conjecture that the entropy is monotonically increasing with $L$ for sign activation
function for $L\geq 1$, while it is decreasing with ReLU activation functions after the initial increase.

\subsection{Numerical Computation of Entropy of Functions}

In this section we provide some numerical examples of the resulting entropy in different cases. The
entropy of $\configvec{S}^{L}$, $\mathcal{H}^{L}$ is computed according to Eq.~(\ref{eq:def_entropy});
for $n=2$, the entropy can be computed exactly by calculating the
orthant probability $\text{Prob}(h_{\gamma}^{L}\geq0)$; for $n>2$,
we can use Monte Carlo method to sample $\configvec{h}^{L}$ (which
is straightforward since it follows a multivariate Gaussian distribution)
and estimate $\mathcal{H}^{L}[P(\configvec{S}^{L})]$ accordingly.
The obtained entropy $\mathcal{H}^{L}$ for neural networks with sign
and ReLU activation functions without bias variables are shown in
the main text.

\subsubsection{The Effect of Bias Variables}

Without bias variables, the sign-networks are rather chaotic and eventually
converge to uniform distribution of all Boolean functions. Introducing
bias variables can change the picture, since the local field $h^{l}_{i} = \sum_{j} W^{l}_{ij} S^{l-1}_{j} + b^{l}_{i}$
will be less sensitive to the input variables.

\begin{figure}
\centering\includegraphics[scale=0.5]{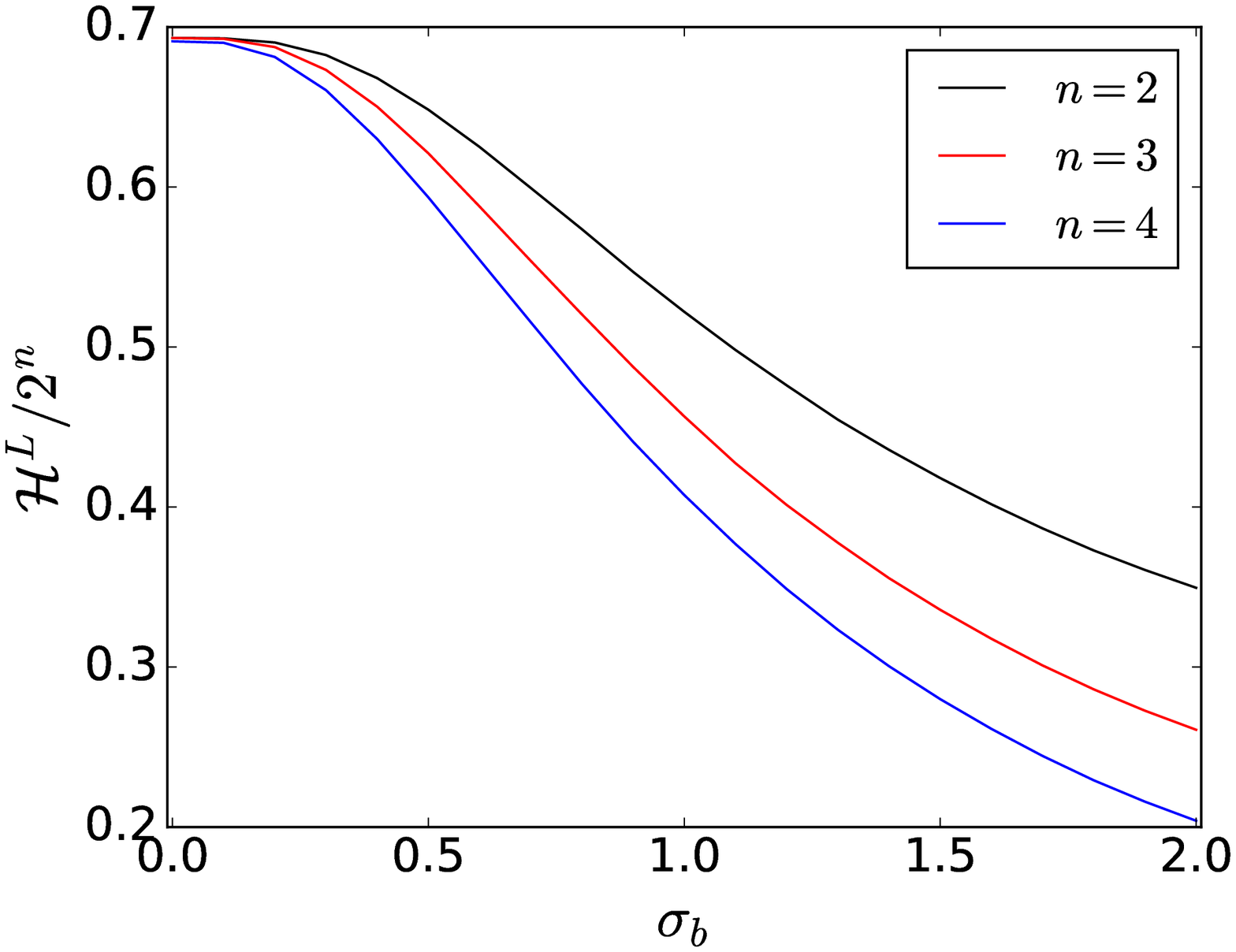}

\caption{Entropy of Boolean functions vs the variance of bias parameters $\sigma_{b}$ in networks with sign activation functions. The variance of weight parameters is
$\sigma_{w}=1$, and the limit $L \to \infty$ is considered.\label{fig:entropy_varb_sb}}
\end{figure}

In the infinite $L$ limit, the fixed point of overlap propagation
is given by (also see Fig.~\ref{fig:iter_map_wb}) 
\begin{equation}
q_{\gamma\gamma'}^{*}=\frac{2}{\pi}\sin^{-1}\bigg(\frac{\sigma_{w}^{2}q_{\gamma\gamma'}^{*}+\sigma_{b}^{2}}{\sigma_{w}^{2}+\sigma_{b}^{2}}\bigg).
\end{equation}
So all the off-diagonal matrix elements of $\configvec{c}^{L}$ will
converge to the same value $\sigma_{w}^{2}q^{*}+\sigma_{b}^{2}$,
while all the diagonal matrix elements are $\sigma_{w}^{2}+\sigma_{b}^{2}$.
The entropy of $\configvec{h}^{L}$ and $\configvec{S}^{L}$ is depicted
in Fig.~\ref{fig:entropy_varb_sb}. The variability of the functions
being computed is decreasing with $\sigma_{b}$.

\section{Generating Functional Analysis of Sparsely Connected Boolean Circuits}

The analysis in neural networks described above applies similarly to the sparsely connected Boolean circuits (we use the term circuit to include both layer-dependent and recurrent discrete Boolean networks and distinguish them from the real variable networks used before). To accommodate the computation with noise, we consider the output of the ($l,i$)-th gate as, 
\begin{eqnarray}
\Spin_{i}^l=\eta_{i}^l\xi_i^l\alpha_i^l
(\Spin_{i_1}^{l-1},\ldots,\Spin_{i_k}^{l-1})\label{def:algorithm}
\end{eqnarray}
where $\eta_{i}^l$ is an independent random variable from the distribution  $\Prob(\eta)=\epsilon\delta_{\eta;-1}\!+\!(1\!-\!\epsilon)\delta_{\eta;1}$ which represents the \emph{dynamic} (annealed) noise and $\xi_i^l$ is an independent random variables from the distribution  $\Prob(\xi)=p\delta_{\xi;-1}+(1-p)\delta_{\xi;1}$ which represents the \emph{quenched} (hard) noise. We note that the annealed noise is \emph{different} for each copy of the system but the quenched noise (and topology) is the \emph{same} in all copies.

The generating functional is 
\begin{eqnarray}
\Gamma[\{\psi_{i,\gamma }^{l}\}]&=&\sum_{\{S_{i,\gamma }^{l}\}}\!\prod_{\gamma=1}^{2^n}\!\Prob(\sitevec{S}^0_\gamma\vert s_1^\gamma,\ldots,s_n^\gamma)\!\prod_{l=1}^L\! \Prob(\sitevec{S}^l_\gamma\vert\sitevec{S}^{l-1}_\gamma)\rme^{-\rmi\sum_{l,i}\psi_{i,\gamma }^{l} S_{i,\gamma }^{l}},\label{def:GF}
\end{eqnarray}
where 
\begin{eqnarray}
\Prob(\sitevec{S}^l_\gamma \vert \sitevec{S}^{l-1}_\gamma )&=&\prod_{i=1}^N\frac{\rme^{\beta S_{i,\gamma }^l\sum_{j_1,\ldots,j_k}^N A_{j_1,\ldots,j_k}^{l,i}\xi_i^l\alpha_i^l(S_{j_1,\gamma }^{l-1},\ldots,S_{j_k,\gamma }^{l-1})}}{2\cosh[\beta\sum_{j_1,\ldots,j_k}^N A_{j_1,\ldots,j_k}^{l,i}\xi_i^l\alpha_i^l(S_{j_1,\gamma }^{l-1},\ldots,S_{j_k,\gamma }^{l-1})]}.\label{eq:layerProb}
\end{eqnarray}
We have averaged out the noise variables and the inverse temperature $\beta$ is related to the noise parameter $\epsilon$ via $\tanh\beta=1-2\epsilon$.

The set of connectivity tensors $\{A_{i_1,\ldots,i_k}^{l,i}\}$,
where $A_{i_1,\ldots,i_k}^{l,i}\in\{0,1\}$, denotes connections in the circuit. The sources of disorder in our model are the random connections, random boundary conditions and random gates. The former two arise in the layered growth process. The basic change in this growth process is the addition of a new gate with probability $\Prob(A_{j_1,\ldots,j_k}^{l,i})=\frac{1}{N^k}\delta_{A_{j_1,\ldots,j_k}^{l,i};1}+(1-\frac{1}{N^k})\delta_{A_{j_1,\ldots,j_k}^{l,i};0}$ of being connected to exactly $k$ gate-outputs of the previous layer $l-1$. This procedure is carried out independently for all gates in the circuit giving rise to the probability distribution 
\begin{eqnarray}
\Prob(\{A_{i_1,\ldots,i_k}^{l,i}\})&=&\frac{1}{Z_A}\prod_{l,i=1}^{L,N}\Bigg [\delta\bigg(1; \sum_{j_1,\ldots,j_k}^N A_{j_1,\ldots,j_k}^{l,i}\bigg)\prod_{i_1,\ldots,i_k}^{N}\!\!\left[\frac{1}{N^k}\delta_{A_{i_1,\ldots,i_k}^{l,i};1}+(1-\frac{1}{N^k})\delta_{A_{i_1,\ldots,i_k}^{l,i};0}\right]\Bigg ], \label{def:connect-disorder}
\end{eqnarray}
where $Z_A$ is a normalization constant. The Kronecker delta function inside the definition (\ref{def:connect-disorder}) enforces the constraint $\sum_{j_1,\ldots,j_k}^N A_{j_1,\ldots,j_k}^{l,i}=1$, i.e. the gate on site $(l, i)$ is mapped to exactly one element from the set of all possible output-indices $\{i_1,\ldots,i_k\}$ from the previous layer.  

Random boundary conditions in the layered growth process are generated by selecting indices to the entries of the input vector $\sitevec{S}^I$ with probability $\frac{1}{\vert \sitevec{S}^I\vert}$, and assigning them to the initial layer $l=0$. 

In addition to the topological disorder, induced by the growth process, we assume that the gate $\alpha_{i}^{l}$ added at each step of the process can be sampled randomly and independently from the set $G$ of $k$-ary Boolean gates. Under this assumption the distribution over gates takes the form 
\begin{eqnarray}
\Prob(\{\alpha_{i}^{l}\})=\prod_{l=1}^{L} \prod_{i=1}^{N} \Prob(\alpha_{i}^{l}),\label{def:gate-disorder}
\end{eqnarray}
where $\Prob(\alpha_{i}^{l})=\sum_{\alpha\in G}p_{\alpha}\delta_{\alpha;\alpha_{i}^{l}}$ with $\sum_{\alpha\in G}p_{\alpha}=1$ and $p_{\alpha}\geq0$. 

In the simplest case as discussed in the main text, a single gate $\alpha$ is used with $\Prob(\alpha_{i}^{l})=\delta_{\alpha;\alpha_{i}^{l}}$. 
The diluted networks generated are random directed networks, as the incoming edges and out-going edges are distinguished. Since in the current analysis the same gate is used for the whole architecture, all in-degrees of nodes are the same, while the randomly selected out-degrees follow a Poisson distribution, hence there is a small probability that a node does not contribute to the next layer nodes

\subsection{Layer-dependent Architectures}
We assume that the system is self-averaging, i.e. any macroscopic quantity which can be computed via Eq.~(\ref{def:GF}) is self-averaging, and compute the disorder-averaged generating functional
\begin{eqnarray}
\overline{\Gamma[\{\psi_{i,\gamma }^{l}\}]}&=&\sum_{\{S_{i,\gamma }^{l}\}}\!\prod_{\gamma =1}^{2^n}\!\Prob(\sitevec{S}^0_\gamma \vert s_1^\gamma,\ldots,s_n^\gamma )\;\rme^{-\rmi\sum_{l,i}\psi_{i,\gamma }^{l} S_{i,\gamma }^{l}}\overline{\prod_{\gamma =1}^{2^n}\prod_{l=1}^L\!
\prod_{i=1}^N\frac{\rme^{\beta S_{i,\gamma }^l  h_{i,\gamma }^{l-1}(\sitevec{S}_\gamma ^{l-1})}}{2\cosh[\beta  h_{i,\gamma }^{l-1}(\sitevec{S}_\gamma ^{l-1})]}},\label{eq:GF1}
\end{eqnarray}
defining the field 
\begin{eqnarray}
h_{i,\gamma }^{l-1}(\sitevec{S}_{\gamma}^{l-1})=\sum_{j_1,\ldots,j_k}^N A_{j_1,\ldots,j_k}^{l,i}\xi_i^l\alpha_i^l(S_{j_1,\gamma }^{l-1},\ldots,S_{j_k,\gamma }^{l-1}).
\end{eqnarray}
Notice that the index convention of the field $H$ is different from the case of fully connected neural networks.

Isolating the fields $\{ h_{i,\gamma }^{l-1}(\sitevec{S}^{l-1})\}$ in Eq.~(\ref{eq:GF1}) via the integral representations of unity 
\begin{eqnarray}
\prod_{\gamma =1}^{2^n}\prod_{l=1}^{L}\prod_{i=1}^{N}\left\{\int\frac{\mathrm d  h_{i,\gamma }^{l-1}\mathrm d  x_{i,\gamma }^{l-1}}{2\pi}\rme^{\rmi  x_{i,\gamma }^{l-1}[ h_{i,\gamma }^{l-1}\!-\! h_{i,\gamma }^{l-1}(\sitevec{S}_{\gamma }^{l-1})]}\right\}\!=\!1\label{def:unity}
\end{eqnarray}
gives us	
\begin{eqnarray}
\overline{\Gamma[\{\psi_{i,\gamma }^{l}\}]}&=&\sum_{\{S_{i,\gamma }^{l}\}}\!\prod_{\gamma =1}^{2^n}\!\Prob(\sitevec{S}^0_\gamma \vert s_1^\gamma,\ldots,s_n^\gamma )\;\rme^{-\rmi\sum_{l,i}\psi_{i,\gamma }^{l} S_{i,\gamma }^{l}}\nonumber \\
&&\times\prod_{\gamma =1}^{2^n}\prod_{l=1}^L\!
\prod_{i=1}^N\frac{\rme^{\beta S_{i,\gamma }^l  h_{i,\gamma }^{l-1}}}{2\cosh\beta  h_{i,\gamma }^{l-1}}\int\frac{\mathrm d  h_{i,\gamma }^{l-1}\mathrm d  x_{i,\gamma }^{l-1}}{2\pi}\rme^{\rmi  x_{i,\gamma }^{l-1}  h_{i,\gamma }^{l-1}}
\nonumber\\
&&\times\overline{\prod_{\gamma =1}^{2^n}\prod_{l=1}^L\!
\prod_{i=1}^N\rme^{-\rmi  x_{i,\gamma }^{l-1} \sum_{j_1,\ldots,j_k}^N A_{j_1,\ldots,j_k}^{l,i}\xi_i^l\alpha_i^l(S_{j_1,\gamma }^{l-1},\ldots,S_{j_k,\gamma }^{l-1})}}.
\label{eq:GF2}
\end{eqnarray}
We compute the disorder averages in the disorder-dependent part of Eq.~(\ref{eq:GF2}) as follows
\begin{eqnarray}
&&\overline{\prod_{\gamma =1}^{2^n}\prod_{l=1}^L\!
\prod_{i=1}^N\prod_{j_1,\ldots,j_k}^N\rme^{-\rmi  x_{i,\gamma }^{l-1}  A_{j_1,\ldots,j_k}^{l,i}\xi_i^l\alpha_i^l(S_{j_1,\gamma }^{l-1},\ldots,S_{j_k,\gamma }^{l-1})}} \nonumber \\
=&&\frac{1}{Z_A}\prod_{l=1}^L\!
\prod_{i=1}^N
\prod_{i_1,\ldots,i_k}^{N}\left\{\sum_{A_{i_1,\ldots,i_k}^{l,i}}\left[\frac{1}{N^k}\delta_{A_{i_1,\ldots,i_k}^{l,i};1}+(1-\frac{1}{N^k})\delta_{A_{i_1,\ldots,i_k}^{l,i};0}\right]\right\}\nonumber \delta\left(1; \sum_{j_1,\ldots,j_k}^N A_{j_1,\ldots,j_k}^{l,i}\right) \nonumber \\
&&\times\sum_{\xi_i^l}\Prob(\xi_i^l)\sum_{\alpha_i^l}\Prob(\alpha_{i}^{l})\prod_{\gamma =1}^{2^n}\prod_{j_1,\ldots,j_k}^N\rme^{-\rmi  x_{i,\gamma }^{l-1}  A_{j_1,\ldots,j_k}^{l,i}\xi_i^l\alpha _i^l(S_{j_1,\gamma }^{l-1},\ldots,S_{j_k,\gamma }^{l-1})}\nonumber\\
=&&\frac{1}{Z_A}\prod_{l=1}^L\!
\prod_{i=1}^N\int_{-\pi}^{\pi}\frac{\mathrm d \omega_i^l}{2\pi}\rme^{\rmi\omega_i^l}\sum_{\xi_i^l}\Prob(\xi_i^l)\sum_{\alpha_i^l}\Prob(\alpha_{i}^{l})\prod_{i_1,\ldots,i_k}^{N}\sum_{A_{i_1,\ldots,i_k}^{l,i}}\left[\frac{1}{N^k}\delta_{A_{i_1,\ldots,i_k}^{l,i};1}+(1-\frac{1}{N^k})\delta_{A_{i_1,\ldots,i_k}^{l,i};0}\right]\nonumber\\
&&\times
\rme^{-\rmi\sum_{\gamma =1}^{2^n}  x_{i,\gamma }^{l-1}  A_{i_1,\ldots,i_k}^{l,i}\xi_i^l\alpha_i^l(S_{i_1,\gamma }^{l-1},\ldots,S_{i_k,\gamma }^{l-1})-\rmi\omega_i^l A_{i_1,\ldots,i_k}^{l,i}}\nonumber
\nonumber\\
=&&\frac{1}{Z_A}\prod_{l=1}^L\!
\prod_{i=1}^N\int_{-\pi}^{\pi}\frac{\mathrm d \omega_i^l}{2\pi}\rme^{\rmi\omega_i^l} \prod_{i_1,\ldots,i_k}^{N}\left\langle\frac{1}{N^k}\rme^{-\rmi\sum_{\gamma=1}^{2^n}  x_{i,\gamma}^{l-1}\xi\alpha(S_{i_1,\gamma}^{l-1},\ldots,S_{i_k,\gamma}^{l-1})-\rmi\omega_i^l}+(1-\frac{1}{N^k})\right\rangle_{\xi,\alpha}\nonumber\\
=&&\frac{1}{Z_A}\prod_{l=1}^L\!\left\{\prod_{i=1}^N\int_{-\pi}^{\pi}\frac{\mathrm d \omega_i^l}{2\pi}\rme^{\rmi\omega_i^l}\right\} \exp\!\left[\frac{1}{N^k}\!\sum_{i,i_1,\ldots,i_k}^{N}\!\left\langle\rme^{-\rmi\sum_{\gamma=1}^{2^n}  x_{i,\gamma}^{l-1}\xi\alpha(S_{i_1,\gamma}^{l-1},\ldots,S_{i_k,\gamma}^{l-1})-\rmi\omega_i^l} -1\right\rangle_{\xi,\alpha}+O(N^{-k+1})\right]
\end{eqnarray} 

Using the result of disorder average in the generating functional Eq.~(\ref{eq:GF2}) gives us
\begin{eqnarray}
\overline{\Gamma}&=&\frac{1}{Z_A}\sum_{\{S_{i,\gamma}^{l}\}}\!\prod_{\gamma=1}^{2^n}\!\Prob(\sitevec{S}^0_\gamma\vert s_1^\gamma,\ldots,s_n^\gamma)\rme^{-\rmi\sum_{l,i}\psi_{i,\gamma}^{l} S_{i,\gamma}^{l}} \nonumber\\
&&\times \left\{ \prod_{l,\gamma,i}\int\frac{\mathrm d h_{i,\gamma}^l\mathrm d  x_{i,\gamma}^{l-1}}{2\pi}
\rme^{\rmi  x_{i,\gamma}^{l-1} h_{i,\gamma}^l} \right\} \rme^{\beta \sum_{l,\gamma,i}S_{i,\gamma}^l  h_{i,\gamma}^{l-1}+\sum_{l,\gamma,i}\log2\cosh\beta  h_{i,\gamma}^{l-1}}
\nonumber\\
&&\times\prod_{l=1}^L\!\left\{\prod_{i=1}^N\int_{-\pi}^{\pi}\frac{\mathrm d \omega_i^l}{2\pi}\rme^{\rmi\omega_i^l}\right\}\exp\!\Big[N\int\rmd\configvec{x}\rmd\omega\frac{1}{N}\sum_{i=1}^{N}\delta(\configvec{x}\!-\!\configvec{x}_i^{l-1})\delta(\omega\!-\!\omega^l_i)\nonumber\\
&&\times\!\!\!\sum_{\configvec{S}_1,..,\configvec{S}_k}\frac{1}{N^k}\!\!\sum_{i_1,\ldots,i_k}^{N}\!\!\delta_{\configvec{S}_1;\configvec{S}_{i_1}^{l-1}}\times\!\cdots\!\times\delta_{\configvec{S}_k;\configvec{S}_{i_k}^{l-1}} \left\langle\rme^{-\rmi\sum_{\gamma=1}^{2^n} x_{\gamma}\xi\alpha(S_{1}^\gamma,\ldots,S_{k}^\gamma)-\rmi\omega} -1\right\rangle_{\xi,\alpha}+O(N^{-k+1})\Big] \label{eq:GF3}
\end{eqnarray}
where we used the definitions $\configvec{x}=(x^1,\ldots,x^{2^n})$ and $\configvec{S}_j=(S_{j}^1,\ldots,S_{j}^{2^n})$, where $j\in\{1,\ldots,k\}$.

In order to achieve the factorization over sites, we insert the following integro-functional representations of unity
\begin{eqnarray}
&&\int\{\mathrm d P^l \mathrm d\hat{P}^l\}\rme^{\rmi N\!\sum_{\configvec{S}}\hat{P}^l(\configvec{S})[P^l(\configvec{S})-\frac{1}{N}\sum_{i=1}^N\delta_{\configvec{S};\configvec{S}_{i}^{l}}]}=1\\
&&\int\{\mathrm d\Omega^l \mathrm d\hat{\Omega}^l\}\rme^{\rmi N\! \int\rmd \configvec{x}\rmd \omega\hat{\Omega}^l(\configvec{x},\omega)[\Omega^l(\configvec{x},\omega)\!-\!\frac{1}{N}\sum_{i=1}^N\delta(\configvec{x}\!-\!\configvec{x}_i^l)\delta(\omega\!-\!\omega^{l+1}_i)]}=1\nonumber
\end{eqnarray}
into the generating functional Eq.~(\ref{eq:GF3}),  which leads to
\begin{eqnarray}
\overline{\Gamma}&=&\frac{1}{Z_A}\sum_{\{S_{i,\gamma}^{l}\}}\!\prod_{\gamma=1}^{2^n}\!\Prob(\sitevec{S}^0_\gamma\vert s_1^\gamma,\ldots,s_n^\gamma)\;\rme^{-\rmi\sum_{l,i}\psi_{i,\gamma}^{l} S_{i,\gamma}^{l}} \nonumber \\
&&\times\left\{\prod_{l,\gamma,i}\int\frac{\mathrm d h_{i,\gamma}^{l-1}\mathrm d  x_{i,\gamma}^{l-1}}{2\pi}\rme^{\rmi  x_{i,\gamma}^{l-1} h_{i,\gamma}^{l-1}}\right\} \rme^{\beta \sum_{l,\gamma,i}S_{i,\gamma}^l  h_{i,\gamma}^{l-1}+\sum_{l,\gamma,i}\log2\cosh\beta  h_{i,\gamma}^{l-1}}
\nonumber\\
&&\times\prod_{l=1}^L\!\left\{\prod_{i=1}^N\int_{-\pi}^{\pi}\frac{\mathrm d \omega_i^l}{2\pi}\rme^{\rmi\omega_i^l}\right\}\int\{\mathrm d\Omega^{l-1} \mathrm d\hat{\Omega}^{l-1}\}\rme^{\rmi N\! \int\rmd \configvec{x}\rmd \omega\hat{\Omega}^{l-1}(\configvec{x},\omega)[\Omega^{l-1}(\configvec{x},\omega)\!-\!\frac{1}{N}\sum_{i=1}^N\delta(\configvec{x}\!-\!\configvec{x}_i^{l-1})\delta(\omega\!-\!\omega^l_i)]}\nonumber\\
&&\times\int\{\mathrm d P^{l-1} \mathrm d\hat{P}^{l-1}\}\rme^{\rmi N\!\sum_{\configvec{S}}\hat{P}^{l-1}(\configvec{S})[P^{l-1}(\configvec{S})-\frac{1}{N}\sum_{i=1}^N\delta_{\configvec{S};\configvec{S}_{i}^{l-1}}]}\nonumber\\
&&\times\exp\!\Big[N\int\rmd\configvec{x}\rmd\omega\Omega^{l-1}(\configvec{x},\omega)\sum_{\configvec{S}_1,..,\configvec{S}_k}P^{l-1}(\configvec{S}_1) \times\cdots\times P^{l-1}(\configvec{S}_k)\nonumber\\
&& \quad \times\left\langle\rme^{-\rmi\sum_{\gamma=1}^{2^n} x_{\gamma}\xi\alpha(S_{1}^\gamma,\ldots,S_{k}^\gamma)-\rmi\omega} -1\right\rangle_{\xi,\alpha}+O(N^{-k+1})\Big]. \label{eq:GF4}
\end{eqnarray}
The objective now is to reduce the above equation to a saddle-point integral. This can be achieved if we define two functionals. The first functional is given by
\begin{eqnarray}
\Psi&=&\!-\!\frac{1}{N}\log Z_A\!+\!\sum_{l=0}^{L-1}\Big\{\int\rmd \configvec{x}\rmd \omega\rmi\hat{\Omega}^{l}(\configvec{x},\omega)\Omega^{l}(\configvec{x},\omega)\!+\! \sum_{\configvec{S}}\rmi\hat{P}^{l}(\configvec{S})P^{l}(\configvec{S})\nonumber \\
&&\!+\!\int\rmd\configvec{x}\rmd\omega\Omega^{l}(\configvec{x},\omega)\sum_{\{\configvec{S}_j\}}\prod_{j=1}^{k}\left\{P^{l}(\configvec{S}_j)\right\}\left\langle\rme^{-\rmi\sum_{\gamma=1}^{2^n} x_{\gamma}\xi\alpha(S_{1}^\gamma,\ldots,S_{k}^\gamma)-\rmi\omega} \!-\!1\right\rangle_{\xi,\alpha}\Big\}\nonumber\\
&&+\frac{1}{N}\sum_{i=1}^N\delta_{m;n_i}\log\sum_{\configvec{S}_i}\int\{\mathrm d \configvec{h}_i\mathrm d \configvec{x}_i\mathrm d\vecomega_i\}\mathcal{M}_m[\configvec{S}_i,\configvec{h}_i\vert\configvec{x}_i, \vecomega_i,\vecPsi_i], \label{def:saddle}
\end{eqnarray}
and the second functional is given by
\begin{eqnarray}
\mathcal{M}_{n_i}[\configvec{S}_i,\configvec{h}_i\vert\configvec{x}_i, \vecomega_i,\vecPsi_i]&=&\prod_{\gamma=1}^{2^n}\left\{ \delta_{S^0_{i,\gamma};S^I_{n_{i}}(s_1^\gamma,\ldots,s_n^\gamma)}\right\}\rme^{-\rmi\sum_{l,\gamma}\psi_{i,\gamma}^{l} S_{i,\gamma}^{l}} \nonumber \\
&&\times\rme^{\sum_{l=0}^{L-1}\sum_{\gamma=1}^{2^n}\{\rmi  x_{i,\gamma}^{l} h_{i,\gamma}^{l}+\beta S_{i,\gamma}^{l+1}  h_{i,\gamma}^{l}+\log2\cosh\beta  h_{i,\gamma}^{l}\}}
\nonumber\\
&&\times\rme^{\sum_{l=0}^{L\!-\!1}\{-\rmi \hat{\Omega}^{l}(\configvec{x_i^{l}},\omega^{l+1}_i)+\rmi\omega_i^{l+1}-\rmi\hat{P}^{l}(\configvec{S}_{i}^{l})\}}, \label{def:M}
\end{eqnarray}
where we have used the definition 
\begin{eqnarray}
\int\{\mathrm d \configvec{h}_i\mathrm d \configvec{x}_i\mathrm d\vecomega_i\}=\prod_{l=0}^{L-1}\left\{\prod_{\gamma=1}^{2^n}\left\{\int\frac{\mathrm d h_{i,\gamma}^{l}\mathrm d  x_{i,\gamma}^{l}}{2\pi}\right\}\int_{-\pi}^{\pi}\frac{\mathrm d \omega_i^{l+1}}{2\pi}\right\}.
\end{eqnarray}

Using these definitions in Eq.~(\ref{eq:GF4}) gives us the desired saddle-point integral
\begin{eqnarray}
\overline{\Gamma}&=&\int\{\mathrm d\configvec{\Omega} \mathrm d\hat{\configvec{\Omega}}\mathrm d \configvec{P} \mathrm d\hat{\configvec{P}}\}\rme^{N\Psi[\configvec{\Omega},\hat{\configvec{\Omega}}, \configvec{P},\hat{\configvec{P}}]}\label{eq:integral}.
\end{eqnarray}
For $N\rightarrow\infty$ and with the generating fields $\{\psi_{i,\gamma}^l\}$ are all being set to zero we obtain
\begin{eqnarray}
\Psi&=&\sum_{l=0}^{L-1}\Bigg\{\rmi\int\rmd \configvec{x}\rmd \omega\hat{\Omega}^{l}(\configvec{x},\omega)\Omega^{l}(\configvec{x},\omega)+ \rmi\sum_{\configvec{S}}\hat{P}^{l}(\configvec{S})P^{l}(\configvec{S})\label{eq:saddle}\\
&& + \int\rmd\configvec{x}\rmd\omega\Omega^{l}(\configvec{x},\omega)\sum_{\{\configvec{S}_j\}}\bigg\{\prod_{j=1}^{k}P^{l}(\configvec{S}_j)\bigg\}\left\langle\rme^{-\rmi\sum_{\gamma=1}^{2^n} x_{\gamma}\xi\alpha(S_{1}^\gamma,\ldots,S_{k}^\gamma)-\rmi\omega}\right\rangle_{\xi,\alpha}\Bigg\}\nonumber\\
&&+\sum_{m}P(m)\log\sum_{\configvec{S}}\int\{\mathrm d \configvec{h}\mathrm d \configvec{x}\mathrm d\vecomega\} \mathcal{M}_m[\configvec{S},\configvec{h}\vert\configvec{x}, \vecomega,0]\nonumber.
\end{eqnarray}
\subsubsection{Saddle-point Problem}
The integral Eq.~(\ref{eq:integral}) is dominated by the extremum of the functional Eq.~(\ref{eq:saddle}). Functional variation of Eq.~(\ref{eq:saddle}) with respect to the integration variables $\{\configvec{\Omega},\hat{\configvec{\Omega}}, \configvec{P},\hat{\configvec{P}}\}$ leads us to four saddle-point equations
\begin{eqnarray}
&&P^{l}(\configvec{S})=\sum_m\Prob(m)\left\langle\delta_{\configvec{S}^l; \configvec{S}}\right\rangle_{\mathcal{M}_{m}},\label{eq:SP1}\\
&&\Probhat^l(\configvec{S})=\frac{\delta}{\delta P^{l}(\configvec{S})}\sum_{\{\configvec{S}_j\}}\prod_{j=1}^{k}\left[P^{l}(\configvec{S}_j)\right]\label{eq:SP2}\\
&&~~~~~~~~~~~\times\int\rmd\configvec{x}\rmd\omega\Omega^{l}(\configvec{x},\omega)\left\langle\rme^{-\rmi\sum_{\gamma=1}^{2^n} x_{\gamma}\xi\alpha(S_{1}^\gamma,\ldots,S_{k}^\gamma)-\rmi\omega}\right\rangle_{\xi,\alpha},\nonumber\\
&&\Omega^{l}(\configvec{x},\omega)=\sum_m\Prob(m)\left\langle\delta(\configvec{x}-\configvec{x}^l)\delta(\omega-\omega^{l+1})\right\rangle_{\mathcal{M}_m},\label{eq:SP3}\\
&&\hat{\Omega}^{l}(\configvec{x},\omega)=\rmi\sum_{\{\configvec{S}_j\}}\prod_{j=1}^{k}\left\{P^{l}(\configvec{S}_j)\right\}\left\langle\rme^{-\rmi\sum_{\gamma=1}^{2^n} x_{\gamma}\xi\alpha(S_{1}^\gamma,\ldots,S_{k}^\gamma)-\rmi\omega}\right\rangle_{\xi,\alpha}.\label{eq:SP4}
\end{eqnarray}
Inserting the result Eq.~(\ref{eq:SP4}) into Eq.~(\ref{def:M}) and integrating continuous variables leads to 
\begin{eqnarray}
&&\int\{\mathrm d \configvec{h}\mathrm d \configvec{x}\mathrm d\vecomega\}\mathcal{M}_m[\configvec{S},\configvec{h}\vert\configvec{x}, \vecomega,0]\label{eq:M}\\
=&&\prod_{\gamma=1}^{2^n}\left\{ \delta_{S^0_{\gamma};S^I_{m}(s_1^\gamma,\ldots,s_n^\gamma)}\right\}\int\{\mathrm d \configvec{h}\mathrm d \configvec{x}\mathrm d\vecomega\}\prod_{l=0}^{L-1}\rme^{\sum_{\gamma=1}^{2^n}\{\rmi  x_{\gamma}^{l} h_{\gamma}^{l}+\beta S_{\gamma}^{l+1}  h_{\gamma}^{l}+\log2\cosh\beta  h_{\gamma}^{l}\}}\;\rme^{-\rmi\hat{P}^{l}(\configvec{S}^{l})}
\nonumber\\
&&\times\exp\Bigg[\sum_{\{\configvec{S}_j\}}\left\{\prod_{j=1}^{k}P^{l}(\configvec{S}_j)\right\} \left\langle\rme^{-\rmi\sum_{\gamma=1}^{2^n} x_{\gamma}^{l}\xi\alpha(S_{1}^\gamma,\ldots,S_{k}^\gamma)-\rmi\omega^{l+1}}\right\rangle_{\xi,\alpha}   +\rmi\omega^{l+1}\Bigg]\nonumber\\
=&&\prod_{\gamma=1}^{2^n}\left\{ \delta_{S^0_{\gamma};S^I_{m}(s_1^\gamma,\ldots,s_n^\gamma)}\right\}\int\{\mathrm d \configvec{h}\mathrm d \configvec{x}\}\prod_{l=0}^{L-1}\rme^{\sum_{\gamma=1}^{2^n}\{\rmi  x_{\gamma}^{l} h_{\gamma}^{l}+\beta S_{\gamma}^{l+1}  h_{\gamma}^{l}+\log2\cosh\beta  h_{\gamma}^{l}\}}\;\rme^{-\rmi\hat{P}^{l}(\configvec{S}^{l})}
\nonumber\\
&&\times\sum_{\{\configvec{S}_j\}}\prod_{j=1}^{k}\left\{P^{l}(\configvec{S}_j)\right\}\left\langle\rme^{-\rmi\sum_{\gamma=1}^{2^n} x_{\gamma}^{l}\xi\alpha(S_{1}^\gamma,\ldots,S_{k}^\gamma)}\right\rangle_{\xi,\alpha}\nonumber\\
=&&\prod_{\gamma=1}^{2^n}\left\{ \delta_{S^0_{\gamma};S^I_{m}(s_1^\gamma,\ldots,s_n^\gamma)}\right\}\prod_{l=0}^{L-1}\sum_{\{\configvec{S}_j\}}\prod_{j=1}^{k}\left[P^{l}(\configvec{S}_j)\right]\left\langle
\prod_{\gamma=1}^{2^n}\frac{\rme^{\beta S_{\gamma}^{l+1} \xi\alpha(S_{1}^\gamma,\ldots,S_{k}^\gamma)     }}{2\cosh\beta \xi\alpha(S_{1}^\gamma,\ldots,S_{k}^\gamma)}
\right\rangle_{\xi,\alpha}\rme^{-\rmi\sum_{l=0}^{L-1}\hat{P}^{l}(\configvec{S}^{l})}.
\nonumber
\end{eqnarray}
Averaging Eq.~(\ref{eq:M}) over the random-indices disorder $m$ gives 
\begin{eqnarray}
\mathrm{Prob}[\configvec{S}^L\leftarrow\ldots\leftarrow\configvec{S}^0]&=&\sum_m P(m)\left\{\prod_{\gamma=1}^{2^n} \delta_{S^0_{\gamma};S^I_{m}(s_1^\gamma,\ldots,s_n^\gamma)}\right\}\label{eq:effPath}\\
&&\times\prod_{l=0}^{L-1}\sum_{\{\configvec{S}_j\}}\left[\prod_{j=1}^{k}P^{l}(\configvec{S}_j)\right]\left\langle
\prod_{\gamma=1}^{2^n}\frac{\rme^{\beta S_{\gamma}^{l+1} \xi\alpha(S_{1}^\gamma,\ldots,S_{k}^\gamma)     }}{2\cosh\beta \xi\alpha(S_{1}^\gamma,\ldots,S_{k}^\gamma)}
\right\rangle_{\xi,\alpha}\nonumber\\
&&\times\rme^{-\rmi\sum_{l=0}^{L-1}\hat{P}^{l}(\configvec{S}^{l})}/\mathrm{Norm_n},
\nonumber
\end{eqnarray}
where $\configvec{S}^l=(S_1^l,\ldots,S_{2^n}^l)$, which is the probability of a path in the space of Boolean functions of $n$ variables. The conjugate order parameter $\hat P^l$ is a constant for all $l\in\{0,\ldots,L-1\}$ which is canceled by a similar constant in the denominator of Eq.~(\ref{eq:effPath}).

From Eq.~(\ref{eq:effPath}) we can easily obtain the probability of a Boolean function on the layer $l+1$
\begin{eqnarray}
P^{l+1}(\configvec{S})
=\sum_{\{\configvec{S}_j\}}\left\{\prod_{j=1}^{k}P^{l}(\configvec{S}_j)\right\}\left\langle
\prod_{\gamma=1}^{2^n}\frac{\rme^{\beta S_{\gamma} \xi\alpha(S_{1}^\gamma,\ldots,S_{k}^\gamma)     }}{2\cosh\beta \xi\alpha(S_{1}^\gamma,\ldots,S_{k}^\gamma)}
\right\rangle_{\xi,\alpha},\label{eq:P}
\end{eqnarray}
where the initial condition is given by 
\begin{eqnarray}
P^{0}(\configvec{S})&=&\sum_m P(m)\prod_{\gamma=1}^{2^n}\left\{ \delta_{S_{\gamma};S^I_{m}(s_1^\gamma,\ldots,s_n^\gamma)}\right\}.
\end{eqnarray}

Eq.~(\ref{eq:P}) is the main result of this section. 
	
\subsection{Recurrent Architectures}
In recurrent Boolean networks, both random connections and random gates do not change from layer to layer but remain fixed. This, for all $l\in\{1,\ldots,L\}$, gives rise to the probabilities
\begin{eqnarray}
\Prob(\{A_{i_1,\ldots,i_k}^{i}\})&=&\frac{1}{Z_A}\prod_{i=1}^{N}\Bigg[\delta\bigg(1; \sum_{j_1,\ldots,j_k}^N A_{j_1,\ldots,j_k}^{i}\bigg) \prod_{i_1,\ldots,i_k}^{N}\left[\frac{1}{N^k}\delta_{A_{i_1,\ldots,i_k}^{i};1}+(1-\frac{1}{N^k})\delta_{A_{i_1,\ldots,i_k}^{i};0}\right]\Bigg],\label{def:recurrent-connect-disorder}
\end{eqnarray}
and 
\begin{eqnarray}
\Prob(\{\alpha_{i}\})=\prod_{i=1}^{N}\Prob(\alpha_{i}).\label{def:recurrent-gate-disorder}
\end{eqnarray}
The differences between the random layer-dependent and random recurrent topologies are generated in the disorder-dependent part of Eq.~(\ref{eq:GF1}). We average out the disorder in Eq.~(\ref{eq:GF1}) as follows
\begin{eqnarray}
&&\overline{\prod_{i=1}^N\prod_{\gamma =1}^{2^n}\prod_{l=1}^L\!
\prod_{j_1,\ldots,j_k}^N\rme^{-\rmi  x_{i,\gamma }^{l-1}  A_{j_1,\ldots,j_k}^{i}\xi_i\alpha_i(S_{j_1,\gamma }^{l-1},\ldots,S_{j_k,\gamma }^{l-1})}}\nonumber \\
=&&\frac{1}{Z_A}
\prod_{i=1}^N
\prod_{i_1,\ldots,i_k}^{N}\left\{\sum_{A_{i_1,\ldots,i_k}^{i}}\left[\frac{1}{N^k}\delta_{A_{i_1,\ldots,i_k}^{i};1}+(1-\frac{1}{N^k})\delta_{A_{i_1,\ldots,i_k}^{i};0}\right]\right\}\delta\bigg(1; \sum_{j_1,\ldots,j_k}^N A_{j_1,\ldots,j_k}^{i}\bigg)\nonumber\\&&\times\sum_{\xi_i}\Prob(\xi_i)\sum_{\alpha_i}\Prob(\alpha_{i})\prod_{j_1,\ldots,j_k}^N\rme^{-\rmi\sum_{l=1}^L\sum_{\gamma =1}^{2^n}  x_{i,\gamma }^{l-1}  A_{j_1,\ldots,j_k}^{i}\xi_i\alpha _i(S_{j_1,\gamma }^{l-1},\ldots,S_{j_k,\gamma }^{l-1})}\nonumber\\
=&&\frac{1}{Z_A}
\prod_{i=1}^N\int_{-\pi}^{\pi}\frac{\mathrm d \omega_i}{2\pi}\rme^{\rmi\omega_i}
\prod_{i_1,\ldots,i_k}^{N}\sum_{A_{i_1,\ldots,i_k}^{i}}\left[\frac{1}{N^k}\delta_{A_{i_1,\ldots,i_k}^{i};1}+(1-\frac{1}{N^k})\delta_{A_{i_1,\ldots,i_k}^{i};0}\right]\nonumber\\
&&\times\left\langle 
\rme^{-\rmi\sum_{l=1}^L\sum_{\gamma =1}^{2^n}  x_{i,\gamma }^{l-1}  A_{i_1,\ldots,i_k}^{i}\xi_i\alpha_i(S_{i_1,\gamma }^{l-1},\ldots,S_{i_k,\gamma }^{l-1})-\rmi\omega_i A_{i_1,\ldots,i_k}^{i}}\right\rangle_{\xi_i,\alpha_i}\nonumber
\nonumber\\
=&&\frac{1}{Z_A} \prod_{i=1}^N\int_{-\pi}^{\pi}\frac{\mathrm d \omega_i}{2\pi}\rme^{\rmi\omega_i}\prod_{i_1,\ldots,i_k}^{N}\nonumber\left\langle\frac{1}{N^k}\rme^{-\rmi\sum_{l=1}^L\sum_{\gamma=1}^{2^n}  x_{i,\gamma}^{l-1}\xi\alpha(S_{i_1,\gamma}^{l-1},\ldots,S_{i_k,\gamma}^{l-1})-\rmi\omega_i}+(1-\frac{1}{N^k})\right\rangle_{\xi,\alpha}\nonumber\\
=&&\frac{1}{Z_A}\!\left\{\prod_{i=1}^N\int_{-\pi}^{\pi}\frac{\mathrm d \omega_i}{2\pi}\rme^{\rmi\omega_i}\right\}\exp\!\left[\frac{1}{N^k}\!\sum_{i,i_1,\ldots,i_k}^{N}\!\left\langle\rme^{-\rmi\sum_{l=1}^L\sum_{\gamma=1}^{2^n}  x_{i,\gamma}^{l-1}\xi\alpha(S_{i_1,\gamma}^{l-1},\ldots,S_{i_k,\gamma}^{l-1})-\rmi\omega_i} -1\right\rangle_{\xi,\alpha}+O(N^{-k+1})\right],\label{eq:aver-recurr}
\end{eqnarray}
Using the above result in the generating functional Eq.~(\ref{eq:GF2}) we obtain
\begin{eqnarray}
\overline{\Gamma}&=&\frac{1}{Z_A}\sum_{\{S_{i,\gamma}^{l}\}}\!\prod_{\gamma=1}^{2^n}\!\Prob(\sitevec{S}^0_\gamma\vert s_1^\gamma,\ldots,s_n^\gamma)\;\rme^{-\rmi\sum_{l,i}\psi_{i,\gamma}^{l} S_{i,\gamma}^{l}}\nonumber \\
&&\times\left\{\prod_{l,\gamma,i}\int\frac{\mathrm d h_{i,\gamma}^l\mathrm d  x_{i,\gamma}^{l}}{2\pi}\rme^{\rmi  x_{i,\gamma}^{l} h_{i,\gamma}^l}\right\}\nonumber \rme^{\beta \sum_{l,\gamma,i}S_{i,\gamma}^l  h_{i,\gamma}^{l-1}+\sum_{l,\gamma,i}\log2\cosh\beta  h_{i,\gamma}^{l-1}}
\nonumber\\
&&\times\left\{\prod_{i=1}^N\int_{-\pi}^{\pi}\frac{\mathrm d \omega_i}{2\pi}\rme^{\rmi\omega_i}\right\}\exp\!\Bigg[ N\int\mathrm d\configvec{x}^{0}\cdots\mathrm d\configvec{x}^{L\!-\!1}\mathrm d\omega\frac{1}{N}\sum_{i=1}^{N}\bigg(\prod_{l=0}^{L-1}\delta(\configvec{x}^{l}-\configvec{x}_i^{l})\bigg)\delta(\omega-\omega_i)\nonumber\\
&&\quad \times\sum_{\{\configvec{S}^{l}_j\}}\frac{1}{N^k}\!\sum_{i_1,\ldots,i_k}^{N}\bigg(\prod_{l=0}^{L-1}\delta_{\configvec{S}^{l}_1;\configvec{S}_{i_1}^{l}}\times\cdots\times\delta_{\configvec{S}^{l}_k;\configvec{S}_{i_k}^{l}}\bigg) \nonumber\\
&&\quad \times\left\langle\rme^{-\rmi\sum_{l=1}^L\sum_{\gamma=1}^{2^n}  x_{\gamma}^{l-1}\xi\alpha(S_{1,\gamma}^{l-1},\ldots,S_{k,\gamma}^{l-1})-\rmi\omega} -1\right\rangle_{\xi,\alpha}+O(N^{-k+1})\Bigg], \label{eq:GF-recurr-1}
\end{eqnarray}
where we have defined the following vectors $\configvec{x}_i^{l}=(x_{i,1}^{l},\ldots,x_{i,2^n}^{l})$, $\configvec{x}^{l}=(x_{1}^{l},\ldots,x_{2^n}^{l})$, $\configvec{S}_{i}^{l}=(S_{i,1}^{l},\ldots,S_{i,2^n}^{l})$ and $\configvec{S}^{l}=(S_{1}^{l},\ldots,S_{2^n}^{l})$.
 
In order to attain the factorization over sites we insert into Eq.~(\ref{eq:GF-recurr-1}) the following functional unity representations
\begin{eqnarray}
&&\int\{\mathrm d P \mathrm d\hat{P}\}\rme^{\rmi N\!\sum_{\{\configvec{S}^l\}}\hat{P}(\{\configvec{S}^l\})[P(\{\configvec{S}^l\})-\frac{1}{N}\sum_{i=1}^N\prod_{l=0}^{L-1}\delta_{\configvec{S}^l;\configvec{S}_{i}^{l}}]}=1,\\
&&\int\{\mathrm d\Omega \mathrm d\hat{\Omega}\}\rme^{\rmi N\! \int\{\rmd \configvec{x}^l\}\rmd \omega\hat{\Omega}(\{\configvec{x}^l\},\omega)[\Omega(\{\configvec{x}^l\},\omega)\!-\!\frac{1}{N}\sum_{i=1}^N\left[\prod_{l=0}^{L-1}\delta(\configvec{x}^l\!-\!\configvec{x}_i^l)\right]\delta(\omega\!-\!\omega_i)]}=1.\nonumber
\end{eqnarray}

Inserting above into the generating functional Eq.~(\ref{eq:GF-recurr-1}) we obtain 
\begin{eqnarray}
\overline{\Gamma}&=&\int\{\mathrm d P \mathrm d\hat{P}\mathrm d\Omega \mathrm d\hat{\Omega}\} \exp N\Bigg[\rmi\sum_{\{\configvec{S}^l\}}\hat{P}(\{\configvec{S}^l\})P(\{\configvec{S}^l\})+\rmi\int\{\rmd \configvec{x}^l\}\rmd \omega\hat{\Omega}(\{\configvec{x}^l\},\omega)\Omega(\{\configvec{x}^l\},\omega)\nonumber\\
&& \quad +\int\{\mathrm d\configvec{x}^{l}\}\mathrm d\omega\Omega(\{\configvec{x}^l\},\omega)\sum_{\{\configvec{S}^{l}_j\}}P(\{\configvec{S}^l_1\})\times\cdots\times P(\{\configvec{S}^l_k\})\times\cdots
\nonumber\\
&& \quad \cdots\times\left\langle\rme^{-\rmi\sum_{l=1}^L\sum_{\gamma=1}^{2^n}  x_{\gamma}^{l-1}\xi\alpha(S_{1,\gamma}^{l-1},\ldots,S_{k,\gamma}^{l-1})-\rmi\omega} -1\right\rangle_{\xi,\alpha}-\frac{1}{N}\log Z_A\Bigg]\nonumber\\
&&\times\sum_{\{S_{i,\gamma}^{l}\}}\!\prod_{\gamma=1}^{2^n}\!\Prob(\sitevec{S}^0_\gamma\vert s_1^\gamma,\ldots,s_n^\gamma)\;\rme^{-\rmi\sum_{l,i}\psi_{i,\gamma}^{l} S_{i,\gamma}^{l}}\nonumber\\
&&\times\left\{\prod_{l,\gamma,i}\int\frac{\mathrm d h_{i,\gamma}^l\mathrm d  x_{i,\gamma}^{l}}{2\pi}\rme^{\rmi  x_{i,\gamma}^{l} h_{i,\gamma}^l}\right\}\rme^{\beta \sum_{l,\gamma,i}S_{i,\gamma}^l  h_{i,\gamma}^{l-1}+\sum_{l,\gamma,i}\log2\cosh\beta  h_{i,\gamma}^{l-1}}
\nonumber\\
&&\times\left\{\prod_{i=1}^N\int_{-\pi}^{\pi}\frac{\mathrm d \omega_i}{2\pi}\rme^{\rmi\omega_i}\right\}\rme^{-\rmi\sum_{i=1}^N\hat{P}(\{\configvec{S}_i^l\})-\rmi\sum_{i=1}^N\hat{\Omega}(\{\configvec{x}_i^l\},\omega_i)}. \label{eq:GF-recurr-2}
\end{eqnarray}
The site-dependent part of the above can be written as 
\begin{eqnarray}
\exp\Bigg[N\frac{1}{N}\sum_{i=1}^N\delta_{m;n_i}\log\sum_{\{\configvec{S}_i^l\}}\int\{\mathrm d \configvec{h}_i^l\mathrm d \configvec{x}_i^l\}\int_{-\pi}^{\pi}\frac{\mathrm d \omega_i}{2\pi}\times\cdots \times \mathcal{M}_m[\{\configvec{S}_i^l\},\{\configvec{h}_i^l\}\vert\{\configvec{x}_i^l\}, \omega_i,\{\vecPsi_i^l\}]\Bigg], \label{def:Mpart}
\end{eqnarray}
where 
\begin{eqnarray}
\mathcal{M}_{n_i}[\{\configvec{S}_i^l\},\{\configvec{h}_i^l\}\vert\{\configvec{x}_i^l\}, \omega_i,\{\vecPsi_i^l\}] &=& \prod_{\gamma=1}^{2^n}\left\{ \delta_{S^0_{i,\gamma};S^I_{n_{i}}(s_1^\gamma,\ldots,s_n^\gamma)}\right\}\rme^{-\rmi\sum_{l,\gamma}\psi_{i,\gamma}^{l} S_{i,\gamma}^{l}} \\
&& \times\rme^{\sum_{l=0}^{L-1}\sum_{\gamma=1}^{2^n}\{\rmi  x_{i,\gamma}^{l} h_{i,\gamma}^{l}+\beta S_{i,\gamma}^{l+1}  h_{i,\gamma}^{l}+\log2\cosh\beta  h_{i,\gamma}^{l}\}}
\nonumber\\
&& \times\rme^{-\rmi \hat{\Omega}(\{\configvec{x}_i^l\},\omega_i)+\rmi\omega_i-\rmi\hat{P}(\{\configvec{S}_i^l\})}, \label{def:M-recurr}
\end{eqnarray}
and we use the definition $\int\{\mathrm d \configvec{h}_i^l\mathrm d \configvec{x}_i^l\}=\prod_{l=0}^{L-1}\prod_{\gamma=1}^{2^n}\int\frac{\mathrm d h_{i,\gamma}^{l}\mathrm d  x_{i,\gamma}^{l}}{2\pi}$.

The definition Eq.~(\ref{def:Mpart}) allows us to express the disorder-averaged generating functional Eq.~(\ref{eq:GF-recurr-2}) as a saddle-point integral 
\begin{eqnarray}
\overline{\Gamma}&=&\int\{\mathrm d P \mathrm d\hat{P}\mathrm d\Omega \mathrm d\hat{\Omega}\}\rme^{N\Psi[\{P,\hat{P},\Omega,\hat{\Omega}\}]}\label{eq:GF-recurr-sp}
\end{eqnarray}
where 
\begin{eqnarray}
\Psi&=&\sum_{\{\configvec{S}^l\}}\rmi\hat{P}(\{\configvec{S}^l\})P(\{\configvec{S}^l\})+\int\{\rmd \configvec{x}^l\}\rmd \omega\rmi\hat{\Omega}(\{\configvec{x}^l\},\omega)\Omega(\{\configvec{x}^l\},\omega)\label{def:saddle-recurr}\\
&&+\int\{\mathrm d\configvec{x}^{l}\}\mathrm d\omega\Omega(\{\configvec{x}^l\},\omega)\sum_{\{\configvec{S}^{l}_j\}}P(\{\configvec{S}^l_1\})\times\cdots\times P(\{\configvec{S}^l_k\})\times\cdots
\nonumber\\
&&\cdots\times\left\langle\rme^{-\rmi\sum_{l=0}^{L\!-\!1}\sum_{\gamma=1}^{2^n}  x_{\gamma}^{l}\xi\alpha(S_{1,\gamma}^{l},\ldots,S_{k,\gamma}^{l})-\rmi\omega} -1\right\rangle_{\xi,\alpha}-\frac{1}{N}\log Z_A\nonumber\\
&&+\sum_{m}P(m)\log\sum_{\{\configvec{S}^l\}}\int\{\mathrm d \configvec{h}^l\mathrm d \configvec{x}^l\}\int_{-\pi}^{\pi}\frac{\mathrm d \omega}{2\pi}\mathcal{M}_m[\{\configvec{S}^l\},\{\configvec{h}^l\}\vert\{\configvec{x}^l\}, \omega,\{0\}].
\nonumber
\end{eqnarray}
with 
\begin{eqnarray}
\mathcal{M}_{m}[\{\configvec{S}^l\},\{\configvec{h}^l\}\vert\{\configvec{x}^l\}, \omega,\{0\}]&=&P_m(\configvec{S}^0)\prod_{l=0}^{L-1}\frac{\rme^{\beta \configvec{S}^{l+1}\cdot\configvec{h}^{l}}}{\prod_\gamma2\cosh\beta  h_{\gamma}^{l}}\rme^{\rmi  \configvec{x}^{l}\cdot \configvec{h}^{l}}
\nonumber\\
&&\times\rme^{-\rmi \hat{\Omega}(\{\configvec{x}^l\},\omega_i)+\rmi\omega-\rmi\hat{P}(\{\configvec{S}^l\})}\nonumber
\end{eqnarray}
where $P_m(\configvec{S}^0)= \prod_{\gamma=1}^{2^n}\left\{ \delta_{S^0_{\gamma};S^I_{m}(s_1^\gamma,\ldots,s_n^\gamma)}\right\}$; we have removed the generating fields $\{\vecPsi_i^l\}=\{0\} $  and assumed the the law of large numbers $P(m)=\lim_{N\rightarrow\infty}\frac{1}{N}\sum_{i=1}^N\delta_{m;n_i}$ holds.
For $N\rightarrow\infty$ the integral Eq.~(\ref{eq:GF-recurr-sp}) is dominated by the extremum points of Eq.~(\ref{def:saddle-recurr}) i.e. $\frac{\delta\Psi}{\delta P}=0$, $\frac{\delta\Psi}{\delta \hat P}=0$, $\frac{\delta\Psi}{\delta \Omega}=0$ and $\frac{\delta\Psi}{\delta \hat\Omega}=0$. 
 
Computing the extremum points of Eq.~(\ref{def:saddle-recurr}) leads to the saddle-point equations
 \begin{eqnarray}
&&P(\{\configvec{S'}^{l}\})=\sum_m\Prob(m)\left\langle\prod_{l=0}^{L-1}\delta_{ \configvec{S'}^l; \configvec{S}^l}\right\rangle_{\mathcal{M}_m}\label{eq:SP1-recurr}\\
&&\hat{P}(\{\configvec{S}^l\})=\rmi\frac{\delta}{\delta P(\{\configvec{S}^l\})} \!\sum_{\{\configvec{S}^{l}_j\}}\prod_{j=1}^k P(\{\configvec{S}^l_j\})\int\{\mathrm d\configvec{x}^{l}\}\mathrm d\omega\Omega(\{\configvec{x}^l\},\omega)\label{eq:SP2-recurr}\\
&&~~~~~~~~~~~\quad\times \left\langle\rme^{-\rmi\sum_{l=0}^{L\!-\!1}\sum_{\gamma=1}^{2^n}  x_{\gamma}^{l}\xi\alpha(S_{1,\gamma}^{l},\ldots,S_{k,\gamma}^{l})-\rmi\omega}\right\rangle_{\xi,\alpha}    \nonumber\\
&&\Omega(\{\configvec{x'}^l\},\omega')=\sum_m\Prob(m)\left\langle\left[\prod_{l=0}^{L-1}\delta(\configvec{x'}^l-\configvec{x}^l)\right]\delta(\omega'-\omega)\right\rangle_{\mathcal{M}_m}\label{eq:SP3-recurr}\\
&&\hat{\Omega}(\{\configvec{x}^l\},\omega)=\rmi\sum_{\{\configvec{S}^{l}_j\}}\prod_{j=1}^k P(\{\configvec{S}^l_j\})\left\langle\rme^{-\rmi\sum_{l=0}^{L\!-\!1}\sum_{\gamma=1}^{2^n}  x_{\gamma}^{l}\xi\alpha(S_{1,\gamma}^{l},\ldots,S_{k,\gamma}^{l})-\rmi\omega}\right\rangle_{\xi,\alpha}, \label{eq:SP4-recurr}
\end{eqnarray} 
where we use the definition 
\begin{eqnarray}
\langle\cdots\rangle_{\mathcal{M}_m}=\frac{\sum_{\{\configvec{S}^l\}}\int\{\mathrm d \configvec{h}^l\mathrm d \configvec{x}^l\}\int_{-\pi}^{\pi}\frac{\mathrm d \omega}{2\pi}\mathcal{M}_m[\{\configvec{S}^l\},\{\configvec{h}^l\}\vert\{\configvec{x}^l\}, \omega,\{0\}]\cdots }{\sum_{\{\configvec{S}^l\}}\int\{\mathrm d \configvec{h}^l\mathrm d \configvec{x}^l\}\int_{-\pi}^{\pi}\frac{\mathrm d \omega}{2\pi}\mathcal{M}_m[\{\configvec{S}^l\},\{\configvec{h}^l\}\vert\{\configvec{x}^l\}, \omega,\{0\}]}   \nonumber
\end{eqnarray}

Solving the saddle point equations leads to the main result
\begin{eqnarray}
P(\configvec{S}^{L},..,\configvec{S}^{0})&=&\sum_m\Prob(m)P_m(\configvec{S}^0)\sum_{\{\configvec{S}^{l}_j\}}\prod_{j=1}^k P(\{\configvec{S}^l_j\}) \left\langle\prod_{l=0}^{L-1}\prod_{\gamma=1}^{2^n}\frac{\rme^{\beta S_\gamma^{l+1}\xi\alpha(S_{1,\gamma}^{l},\ldots,S_{k,\gamma}^{l})}}{2\cosh\beta  \xi\alpha(S_{1,\gamma}^{l},\ldots,S_{k,\gamma}^{l})}\right\rangle_{\xi,\alpha}
\end{eqnarray} 

We can use this result to generate: (i) single-layer observables
\begin{eqnarray}
P(\configvec{S}^{L})=\sum_{\configvec{S}^{L-1}_j}\left\{\prod_{j=1}^k P(\configvec{S}^{L\!-\!1}_j)\right\}\left\langle\prod_{\gamma=1}^{2^n}\frac{\rme^{\beta S_\gamma^{L}\xi\alpha(S_{1,\gamma}^{L\!-\!1},\ldots,S_{k,\gamma}^{L\!-\!1})}}{2\cosh\beta  \xi\alpha(S_{1,\gamma}^{L\!-\!1},\ldots,S_{k,\gamma}^{L\!-\!1})}\right\rangle_{\xi,\alpha},
\end{eqnarray} 
(ii) two-layer oservables
\begin{eqnarray}
P(\configvec{S}^{L},\configvec{S}^{L^\prime})\!=\!\!\! \sum_{\{\configvec{S}^{L\!-\!1}_j,\configvec{S}^{L^\prime\!-\!1}_j\}}\prod_{j=1}^k P(\configvec{S}^{L\!-\!1}_j,\configvec{S}^{L^\prime\!-\!1}_j) \Bigg\langle\prod_{\gamma=1}^{2^n}\frac{\rme^{\beta S_\gamma^{L}\xi\alpha(S_{1,\gamma}^{L\!-\!1},\ldots,S_{k,\gamma}^{L\!-\!1})}}{2\cosh\beta  \xi\alpha(S_{1,\gamma}^{L\!-\!1},\ldots,S_{k,\gamma}^{L\!-\!1})}\frac{\rme^{\beta S_\gamma^{L^\prime}\xi\alpha(S_{1,\gamma}^{L^\prime\!-\!1},\ldots,S_{k,\gamma}^{L^\prime\!-\!1})}}{2\cosh\beta  \xi\alpha(S_{1,\gamma}^{L^\prime\!-\!1},\ldots,S_{k,\gamma}^{L^\prime\!-\!1})}\Bigg\rangle_{\xi,\alpha}.
\end{eqnarray}

According to (i) the results for the layered growth process Eq.~(\ref{eq:P}) also hold for the recurrent topology.
	
The layers can be viewed as time steps of a dynamical system and a Boolean function $\configvec{f}$, represented by a binary string of length $2^n$, as a molecule of gas with at most $2^{2^n}$ different types of molecules. Molecules of this gas participate in (random) $k$-body collisions specified by the gates. The result of a $k$-body  collision is either a new molecule, i.e. new Boolean function, or one of the molecules involved in the collision. For an infinite number of molecules (or a large finite number of molecules and short times) this  process is described by Eq.~(8) of the main text. Depending on the type of the gate used, i.e. the nature of collision, the system exhibits different behaviors as investigated below.

\section{Some Results of Boolean Circuits}
\subsection{Savicky's Growth Process}

In this section we will use Eq.~(\ref{eq:P}) to study the Boolean functions generated in the layered variant~\cite{Mozeika2010} of the Savicky's growth process~\cite{Savicky1990}. 

But first we will establish the equivalence of these two growth processes. Assuming that the formulae in our growth process are constructed using only a single gate $\alpha$ and that it has no noise, we obtain
\begin{eqnarray}
P^{l\!+\!1}(\configvec{S})
&=&\sum_{\{\configvec{S}_j\}}\left[\prod_{j=1}^{k}P^{l}(\configvec{S}_j)\right]
\prod_{\gamma=1}^{2^n}\delta[S^\gamma;\alpha(S_{1}^\gamma,\ldots,S_{k}^\gamma)], \label{eq:PnoNoise}
\end{eqnarray}
where $\configvec{S}\in\{-1,1\}^{2^n}$.

This result is in agreement with~\cite{Savicky1990}, which establishes a formal equivalence of the layered network and Savicky's formula-growth process.

From~\cite{Mozeika2010} and~\cite{Savicky1990, Brodsky2005} we know that the growth process can converge to a single Boolean function or uniform distribution over some set of all Boolean functions.

\subsubsection{Single Boolean Function}

In the case where the process converges to a single Boolean function $\configvec{f}$ we have $P^{\infty}(\configvec{S})=\prod_{\gamma=1}^{2^n}\delta[S^\gamma;f^\gamma]$, where we represent $\configvec{f}$ as a binary string of length $2^n$.

Inserting above into Eq.~(\ref{eq:PnoNoise}) we obtain 
 \begin{eqnarray}
\prod_{\gamma=1}^{2^n}\delta[S^{\gamma};f^\gamma]&=&\sum_{\{\configvec{S}_j\}}\prod_{j=1}^{k}\left[\prod_{\gamma=1}^{2^n}\delta[S_j^{\gamma};f^\gamma]\right]
\prod_{\gamma=1}^{2^n}\delta[S^{\gamma};\alpha(S_{1}^\gamma,\ldots,S_{k}^\gamma)]\nonumber\\
&=&\prod_{\gamma=1}^{2^n}\delta[S^{\gamma};\alpha(f^\gamma,\ldots,f^\gamma)].
\end{eqnarray}

Thus, when the process converges to a single Boolean function $f$ this implies that $f^\gamma=\alpha(f^\gamma,\ldots,f^\gamma)$ and the number of fixed points of Eq.~(\ref{eq:PnoNoise}) is equal to $2^{2^n}$, i.e. all Boolean functions of $n$ variables.

In order to find out, for a given gate $\alpha$ which satisfies the property $S=\alpha(S,\ldots,S)$, to which Boolean function the growth process converges to (if at all) we have to study the evolution of the probability $P^{l}(\configvec{S})$ which depends on the input set $S^I$ via $P^{0}(\configvec{S})$.

The (Shannon) entropy of a Boolean function $f$ on layer $l$ is defined as $h^l=\!-\!\sum_{\configvec{S}}P^{l}(\configvec{S})\log P^{l}(\configvec{S})$, where we use the convention $P^{l}(\configvec{S})\log P^{l}(\configvec{S})=0$ when $P^{l}(\configvec{S})=0$. 
For $\configvec{S}\in\{-1,1\}^{2^n}$, we have $0\leq h^l\leq2^n\log n$ and $h^l\leq h^l_1\!+\!\cdots\!+\!h^l_{2^n}$, where $h^l_\gamma=\!-\!\sum_{S}P^{l}_\gamma(S)\log P^{l}_\gamma(S)$, which implies $0\leq h^l\leq h^l_1\!+\!\cdots\!+\!h^l_{2^n}$. 

The marginals $P^{l}_\gamma(S)=\frac{1}{2}[1+S m_\gamma^l]$ of (\ref{eq:PnoNoise}) can be computed from the equation

\begin{eqnarray}
m_\gamma^{l+1}=\sum_{\{S_j\}}\prod_{j=1}^{k}\left[\frac{1+ S_j m_\gamma^{l}}{2}\right]\alpha(S_1,\ldots,S_k),\label{eq:m}
\end{eqnarray}
where $\gamma=1,..,2^n$ and $m_\gamma^{0}=\frac{1}{\vert S^I\vert}\sum_{S\in S^I_\gamma(\sitevec{s})} S$. 
 
We notice that for $l\rightarrow\infty$: $h^l_1\!+\!\cdots\!+\!h^l_{2^n}=0$ only when $m_\gamma^{l}\in\{-1,1\}$ for all $\gamma\in\{1,..,2^n\}$.

Let $m_\gamma^{l}=S$, where $S\in\{-1,1\}$, and compute RHS of Eq.~(\ref{eq:m}). The result is given by $\alpha(S,\ldots,S)$. The convergence of the process to a single Boolean function implies that $S=\alpha(S,\ldots,S)$ which consequently implies that $m_\gamma^{l}=\pm1$ are the fixed points of the dynamics Eq.~(\ref{eq:m}).

Let $B_-$ and $B_+$ be basins of attraction of equation of the fixed points $-1$ and $+1$ respectively. For an arbitrary Boolean function $f:\{-1,1\}^n\rightarrow\{-1,1\}$, define two sets of inputs $s_{\pm}$ for which $f(s_1,..,s_n)=\pm1$ then the process Eq.~(\ref{eq:PnoNoise}) converges to the Boolean function $f$ when for all $(s_1,..,s_n)\in s_{\pm}$, $m^{0}=\frac{1}{\vert S^I\vert}\sum_{S\in S^I(\sitevec{s})} S\in B_{\pm}$. 

\begin{figure}
    \centering
    \includegraphics[scale=0.4]{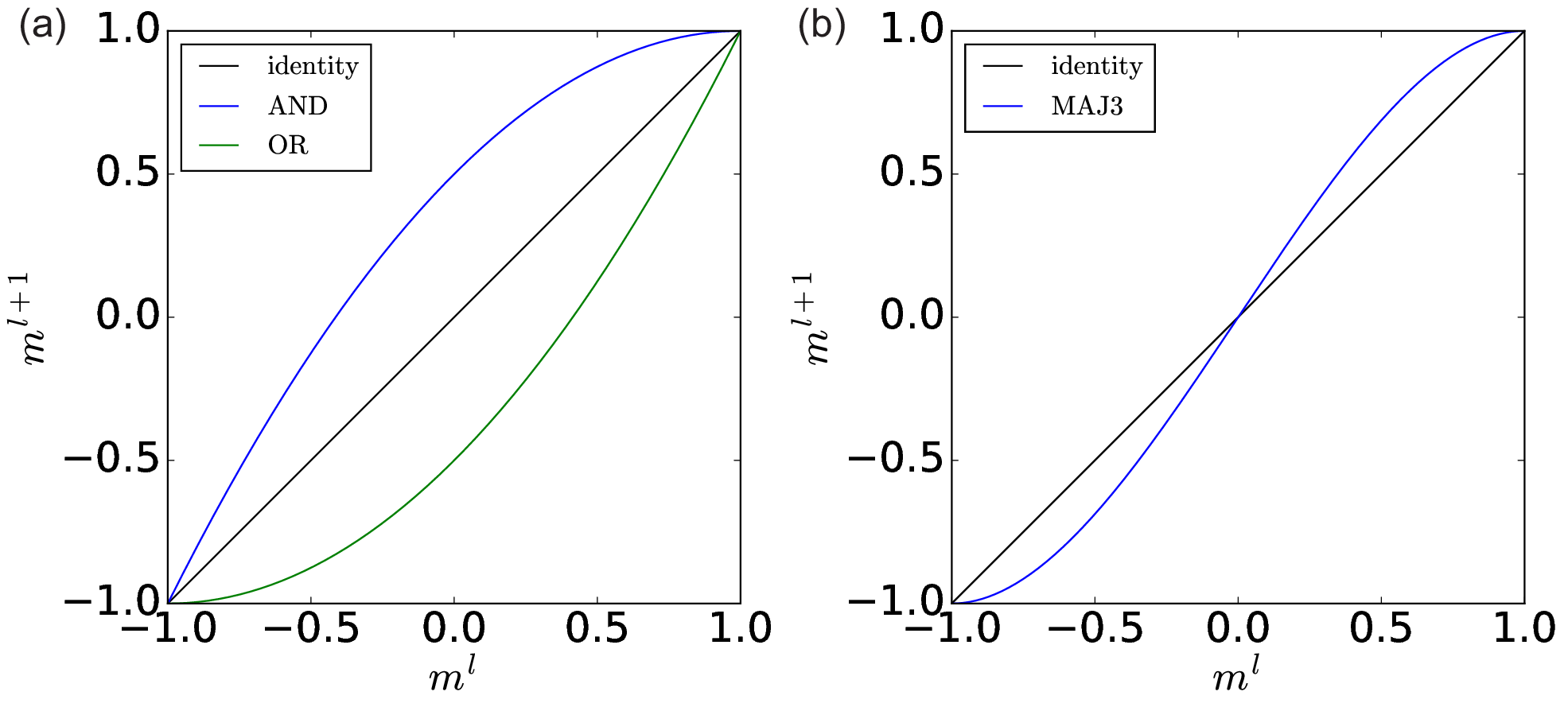}
    \caption{Iteration maps of Eq.~(\ref{eq:m}). (a) AND and OR gate Boolean circuits; $m^{*}=\pm 1$ are the only fixed points of the iteration. (b) MAJ3 gate Boolean circuits; $m^{*}=\pm 1$ are stable fixed points, while $m^{*}=0$ is an unstable fixed point.}
    \label{fig:nextm_vs_m}
\end{figure}

An example of iteration maps of the Eq.~(\ref{eq:m}) for AND and OR gates is plotted in Fig.~(\ref{fig:nextm_vs_m})(a). For these gates, $m_{\gamma}^{l}$ always converges to the fixed points $m^{*} = \pm 1$ as the number of layers grows, which suggests that the corresponding Boolean circuits compute a single Boolean function in the large depth limit. For MAJ3 gate where the iteration map is plotted in Fig.~(\ref{fig:nextm_vs_m})(b), if we consider the input vector $\sitevec{S}^{I} = (s_1, ..., s_n)$, where $n$ is odd, then $m^{0}\neq 0$ and $m_{\gamma}^{l}$ converges to the fixed points $m^{*} = \pm 1$ as well, i.e a single Boolean function is computed in the large depth limit.

Although the analysis is based on the limit of $N \to \infty$ while keeping $n$ finite, we expect this convergence property to hold in the case where $n$ and $N$ are of the same order (e.g., the AND gate loses information of the previous layer and typically contracts the function space), as long as the conditions discussed in this section are satisfied.

\subsubsection{Uniform Distribution over All Boolean Functions}  

In the case when the growth process converges to a uniform distribution over all Boolean functions we have $P^{\infty}(\mathbf{S})=\frac{1}{2^{2^n}}$.

Inserting this distribution into Eq.~(\ref{eq:PnoNoise}) we obtain
\begin{eqnarray}
\frac{1}{2^{2^n}}&=&\sum_{\{\mathbf{S}_j\}}\prod_{j=1}^{k}\left[\frac{1}{2^{2^n}}\right]
\prod_{\gamma=1}^{2^n}\delta[S^{\gamma};\alpha(S_{1}^\gamma,\ldots,S_{k}^\gamma)]\\
&=&\frac{1}{2^{k 2^n}}
\prod_{\gamma=1}^{2^n}\sum_{S_1,\ldots,S_k}\delta[S^{\gamma};\alpha(S_{1},\ldots,S_{k})]\nonumber\\
&=&\frac{1}{2^{k 2^n}}
\prod_{\gamma=1}^{2^n}\left[2^{k-1}+\frac{S^{\gamma}}{2}\sum_{S_1,\ldots,S_k}\alpha(S_{1},\ldots,S_{k})\right]\nonumber
\end{eqnarray}
This equality holds when the gate $\alpha$ is balanced
\begin{equation}
    \sum_{S_1,\ldots,S_k}\alpha(S_{1},\ldots,S_{k})=0. \label{eq:gate_blanced}
\end{equation}

To show that the process converges to $P^{\infty}(\mathbf{S})=\frac{1}{2^{2^n}}$, for a given balanced and nonlinear~\cite{Savicky1990} gate $\alpha$, and initial conditions given by $P^0$, we have to study the dynamics of $P^l$ in general which is beyond the scope of the current study.
Note that in the context of Boolean circuits, the variables are always Boolean and linearity is defined in the finite filed $GF(2)$~\cite{Savicky1990, Brodsky2005}. The requirement for gate $\alpha$ to be nonlinear is due to the fact that any composition of linear Boolean gates is also linear, which implies that only linear Boolean functions can be represented in Boolean circuits using linear gate $\alpha$. Therefore a nonlinear gate is required in order to generate all Boolean functions~\cite{Savicky1990}.

As an example, the MAJ3 gate $\alpha(S_1, S_2, S_3) = \mathrm{sgn}(S_1 + S_2 + S_3)$ is balanced and nonlinear. If the balanced input $\sitevec{S}^{I} = (\sitevec{s}, -\sitevec{s}, 1, -1)$ is used, then $m_{\gamma}^{0} = 0$, which implies that $m_{\gamma}^{l}=0, \forall l\geq 1$. Numerical evidence in the main text shows that (at least for small values of $n$), the entropy of Boolean functions $\mathcal{H}^{L}$ increases monotonically as the number of layers grows and converges to its maximum value $2^{n} \log 2$. It implies that the variability of the Boolean functions computed is increasing and the machines converge to a uniform distribution on all Boolean functions in the large depth limit, which is consistent with the findings in \cite{Savicky1990}.

\begin{figure}
    \centering
    \includegraphics[scale=1.5]{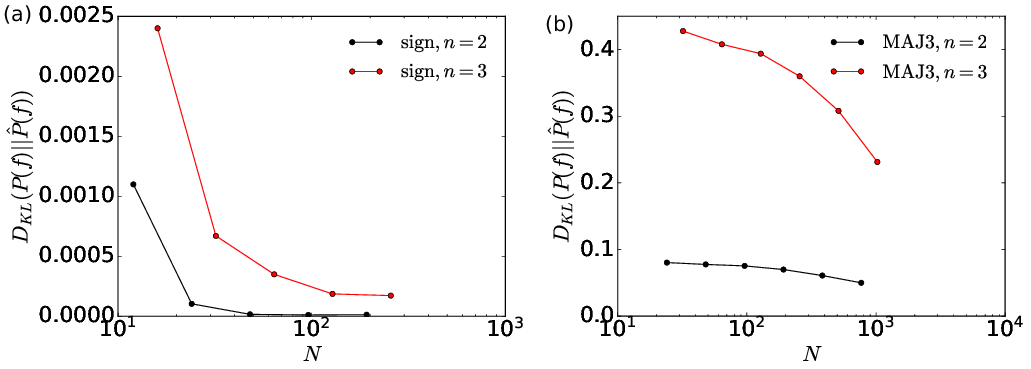}
    \caption{Kullback-Leibler (KL) divergence between the theoretical prediction of the function distribution $P^L(\configvec{f})$ and its estimation $\hat{P}^L(\configvec{f})$ based on finite size simulation (computed by $10^5$ realizations of random networks). In both (a) and (b), $P^L(\configvec{f})$ is close to a uniform distribution. (a) Neural networks based on sign activation function and input vector $\sitevec{S}^{I}=(\sitevec{s}, 1)$. The network depth is $L=8$. (b) Boolean circuits based on MAJ3 gate and input vector $\sitevec{S}^{I} = (\sitevec{s}, -\sitevec{s}, 1, -1)$. The network depth is $L=10$.}
    \label{fig:KL_sign_maj3}
\end{figure}

Unlike the cases where the circuits converge to a single Boolean function for $n$ and $N$ of the same order, here $N$ has to be much larger than $n$; otherwise some intricate Boolean functions may not be computed. This becomes more prominent in sparse Boolean circuits in comparison to densely-connected neural networks. 
Compared to the perceptron (the basic element of neural networks) with variable weight parameters, the Boolean gate is fixed in the Boolean circuit case and the complexity of the computation is realized through the random topologies. Therefore, we expect that the width $N$ in sparse Boolean circuits has to be much larger than the width in neural networks in order to compute complex functions, as illustrated in Fig.~\ref{fig:KL_sign_maj3}.

\section{Identity matrix plus anti-diagonal matrix\label{sec:matrix_Am}}

Consider the $M\times M$ matrix $A_{M}(\kappa)$ with matrix element
\begin{equation}
A_{M}(\kappa)_{ij}=\delta_{ij}+\kappa\delta_{i+j,M+1}.
\end{equation}
For instance,

\begin{equation}
A_{4}(\kappa)=\left(\begin{array}{cccc}
1 & 0 & 0 & \kappa\\
0 & 1 & \kappa & 0\\
0 & \kappa & 1 & 0\\
\kappa & 0 & 0 & 1
\end{array}\right).
\end{equation}
It can be shown that (proof by Laplace expansion and induction) 
\begin{align}
\det\big[A_{M}(\kappa)-\lambda I\big] & =[(1-\lambda)^{2}-\kappa^{2}]^{M/2},\\
\det A_{M}(\kappa) & =(1-\kappa^{2})^{M/2},\\
A_{M}(\kappa)^{-1} & =\frac{1}{1-\kappa^{2}}A_{M}(-\kappa),
\end{align}
which suggests that $\lambda=1\pm\kappa$ are the eigenvalues of $A_{M}(\kappa)$,
each of which has multiplicity of $\frac{M}{2}$. The matrix $A_{M}(\kappa)$
is singular when $\kappa=\pm1$.

\bibliography{ref_sm}